\pdfoutput=1

\documentclass[
12pt,
openany,
english,
onehalfspacing,
liststotoc,
toctotoc,
parskip,
headsepline,
chapterinoneline,
consistentlayout,
]{MastersDoctoralThesis}

\usepackage[utf8]{inputenc}
\usepackage[T1]{fontenc}

\usepackage{mathpazo}

\usepackage[round]{natbib}
\usepackage[autostyle=true]{csquotes}

\usepackage{breakcites}
\usepackage{neuralnetwork}
\usepackage{nameref}
\usepackage{bm}
\usepackage{epigraph}
\usepackage{amsmath}
\usepackage{enumitem}
\usepackage{caption}
\usepackage{microtype}
\usepackage{bookmark}

\thesistitle{Explaining Natural Language\\ Processing Classifiers with Occlusion\\ and Language Modeling}
\supervisor{Prof. Dr. David \textsc{Schlangen}\\
Prof. Dr. Manfred \textsc{Stede}}
\degree{Master of Science}
\author{David \textsc{Harbecke}}

\keywords{}
\university{University of Potsdam}
\program{Cognitive Systems: Language, Learning and Reasoning}
\faculty{Faculty of Human Sciences}

\AtBeginDocument{
\hypersetup{pdftitle=\ttitle}
\hypersetup{pdfauthor=\authorname}
\hypersetup{pdfkeywords=\keywordnames}
}

\begin{document}
\newcommand{\x}[2]{$x_#2$}
\newcommand{\y}[2]{$\hat{y}_#2$}
\newcommand{\hfirst}[2]{\small $h^{1}_#2$}
\newcommand{\hsecond}[2]{\small $h^{2}_#2$}

\newcommand{\insertneuralnetwork}{
\begin{figure}[tb!]
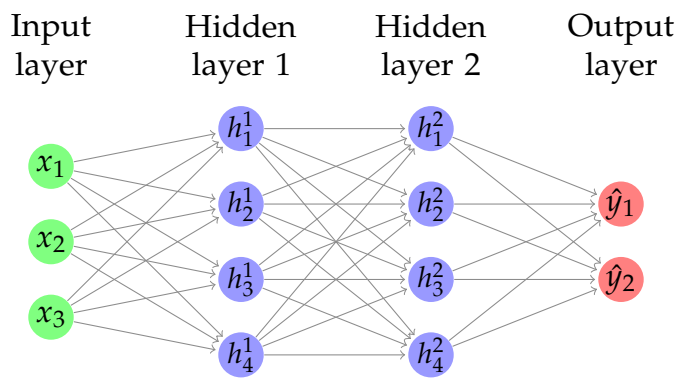

    \centering
    \begin{neuralnetwork}[height=4]
        \inputlayer[count=3, bias=false, title=Input\\layer, text=\x]
        \hiddenlayer[count=4, bias=false, title=Hidden\\layer 1, text=\hfirst] \linklayers
        \hiddenlayer[count=4, bias=false, title=Hidden\\layer 2, text=\hsecond] \linklayers
        \outputlayer[count=2, title=Output\\layer, text=\y] \linklayers
    \end{neuralnetwork}
    \caption[Schematic view of a neural network]
    {Schematic view of a feed-forward neural network with two hidden layers.
    Each node displays a neuron, the arrows between nodes represent the weights.
    The biases and activation functions are not depicted.}
    \label{fig:neural_network}
\end{figure}
}

\newcommand{\insertartificialtable}{
  \begin{figure}[t!]
    \resizebox{\columnwidth}{!}{
    \centering
    \begin{tabular}{lll}
        Explanation 1: &\colorbox[RGB]{255,43,43}{\strut good} \colorbox[RGB]{255,224,224}{\strut film} \colorbox[RGB]{255,254,254}{\strut ,} \colorbox[RGB]{255,178,178}{\strut but} \colorbox[RGB]{193,193,255}{\strut very} \colorbox[RGB]{173,173,255}{\strut glum} \colorbox[RGB]{255,254,254}{\strut .} & (positive sentiment) \\
        &\colorbox[RGB]{255,43,43}{\strut good} \colorbox[RGB]{255,224,224}{\strut film} \colorbox[RGB]{255,254,254}{\strut ,} \colorbox[RGB]{255,178,178}{\strut but} \colorbox[RGB]{193,193,255}{\strut very} \colorbox[RGB]{173,173,255}{\strut glum} \colorbox[RGB]{255,254,254}{\strut .} & (negative sentiment) \bigskip \\
        
        Explanation 2: &\colorbox[RGB]{255,43,43}{\strut good} \colorbox[RGB]{255,224,224}{\strut film} \colorbox[RGB]{255,254,254}{\strut ,} \colorbox[RGB]{255,178,178}{\strut but} \colorbox[RGB]{193,193,255}{\strut very} \colorbox[RGB]{173,173,255}{\strut glum} \colorbox[RGB]{255,254,254}{\strut .} & (positive sentiment) \\
        &\colorbox[RGB]{43,43,255}{\strut good} \colorbox[RGB]{224,224,255}{\strut film} \colorbox[RGB]{254,254,255}{\strut ,} \colorbox[RGB]{178,178,255}{\strut but} \colorbox[RGB]{255,193,193}{\strut very} \colorbox[RGB]{255,173,173}{\strut glum} \colorbox[RGB]{254,254,255}{\strut .} & (negative sentiment)
    \end{tabular}
    }
    \caption[Artificial explanations]{Artificial explanations from two methods. We are investigating the explanation of a sentiment classifier for both positive and negative sentiment. Red indicates a feature supporting the prediction, blue indicates a detraction. The first explanations do not satisfy \emph{Class Zero-Sum} as the explanations are equal for both classes. It is unclear which token actually contributed to which classes. The second explanations do satisfy \emph{Class Zero-Sum}. It is identifiable which token contributed to which class and by how much.}
    \label{fig:axiom_examples}
\end{figure}
}

\newcommand{\insertexampletable}{
 \begin{table*}[hb!]
  \centering
  \begin{tabular}{l|l|l}
    Method&Relevances&Max. value \\ \midrule
    \textbf{OLM}&\colorbox[RGB]{255,91,91}{\strut good} \colorbox[RGB]{255,0,0}{\strut film} \colorbox[RGB]{255,221,221}{\strut ,} \colorbox[RGB]{253,253,255}{\strut but} \colorbox[RGB]{255,190,190}{\strut very} \colorbox[RGB]{255,90,90}{\strut glum} \colorbox[RGB]{255,254,254}{\strut .} &0.57\\
    \textbf{OLM-S}&\colorbox[RGB]{255,43,43}{\strut good} \colorbox[RGB]{255,26,26}{\strut film} \colorbox[RGB]{255,127,127}{\strut ,} \colorbox[RGB]{255,251,251}{\strut but} \colorbox[RGB]{255,89,89}{\strut very} \colorbox[RGB]{255,0,0}{\strut glum} \colorbox[RGB]{255,255,255}{\strut .} &0.48\\
    \midrule
    Delete&\colorbox[RGB]{255,0,0}{\strut good} \colorbox[RGB]{255,235,235}{\strut film} \colorbox[RGB]{255,253,253}{\strut ,} \colorbox[RGB]{250,250,255}{\strut but} \colorbox[RGB]{255,250,250}{\strut very} \colorbox[RGB]{250,250,255}{\strut glum} \colorbox[RGB]{251,251,255}{\strut .} &0.98\\
    UNK&\colorbox[RGB]{255,0,0}{\strut good} \colorbox[RGB]{255,158,158}{\strut film} \colorbox[RGB]{255,229,229}{\strut ,} \colorbox[RGB]{250,250,255}{\strut but} \colorbox[RGB]{250,250,255}{\strut very} \colorbox[RGB]{250,250,255}{\strut glum} \colorbox[RGB]{250,250,255}{\strut .} &0.98\\
    \midrule
    Sensitivity Analysis&\colorbox[RGB]{255,184,184}{\strut good} \colorbox[RGB]{255,160,160}{\strut film} \colorbox[RGB]{255,202,202}{\strut ,} \colorbox[RGB]{255,186,186}{\strut but} \colorbox[RGB]{255,190,190}{\strut very} \colorbox[RGB]{255,0,0}{\strut glum} \colorbox[RGB]{255,211,211}{\strut .} &35\\
    Gradient*Input&\colorbox[RGB]{255,31,31}{\strut good} \colorbox[RGB]{255,0,0}{\strut film} \colorbox[RGB]{255,5,5}{\strut ,} \colorbox[RGB]{214,214,255}{\strut but} \colorbox[RGB]{27,27,255}{\strut very} \colorbox[RGB]{101,101,255}{\strut glum} \colorbox[RGB]{255,236,236}{\strut .} &0.041\\
    Integrated Gradients&\colorbox[RGB]{255,135,135}{\strut good} \colorbox[RGB]{255,229,229}{\strut film} \colorbox[RGB]{255,200,200}{\strut ,} \colorbox[RGB]{251,251,255}{\strut but} \colorbox[RGB]{214,214,255}{\strut very} \colorbox[RGB]{0,0,255}{\strut glum} \colorbox[RGB]{255,224,224}{\strut .} &0.96\\
  \end{tabular}
  \caption[Example explanations for SST-2]{Example explanations for SST-2.
  Explanations are shown for positive sentiment which the input is correctly predicted as.
  Positive relevances are displayed in red, negative relevances are displayed in blue.
  Color intensity is normalized for every explanation method and is proportional to absolute value of relevance.
  The last column gives the maximum of the absolute relevances.
  \emph{OLM} and \emph{OLM-S} give relevance or sensitivity to both clauses.
  The other occlusion-based methods give almost all relevance to \enquote{good}.
  Gradient-based methods give most relevance to the second clause.
  Resamples that \emph{OLM} used multiple times can be found in Table \ref{tab:sampling_examples}.
  \emph{OLM} gives positive relevance to \enquote{glum} because some alternatives are predicted with a much lower probability for positive sentiment.
  }
  \label{tab:example_explanations}
 \end{table*}
}

\newcommand{\insertsamplingtable}{
\begin{table*}[p]
  \resizebox{\columnwidth}{!}{%
  \centering
  \begin{tabular}{|l|l|ll|l|l|l}
    \textbf{good} & \textbf{film} & \multicolumn{1}{l|}{\textbf{,} } & \textbf{but} & \textbf{very} & \textbf{glum} & \multicolumn{1}{l|}{\textbf{.} } \\ 
    \midrule
    good & looking & \multicolumn{1}{l|}{,} & but & not & bad & \multicolumn{1}{l|}{.} \\
    (34, 0.98) & (11, 0.96) & \multicolumn{1}{l|}{(84, 0.98)} & (87, 0.98) & (22, 1) & (26, 0.003) & \multicolumn{1}{l|}{(100, 0.98)} \\ 
    \hline
    nice & news & \multicolumn{1}{l|}{art} & and & still & short & \\
    (10, 0.46) & (5, 0.018) & \multicolumn{1}{l|}{(2, 0.79)} & (3, 1) & (10, 0.87) & (11, 1) & \\ 
    \cline{1-6}
    great & idea & \multicolumn{1}{l|}{quality} & not & very & old & \\
    (3, 0.41) & (4, 0.0011) & \multicolumn{1}{l|}{(2, 1)} & (2, 1) & (10, 0.98) & (5, 1) & \\ 
    \cline{1-6}
    fine & taste & \multicolumn{1}{l|}{} & though & also & thin & \\
    (3, 0.27) & (4, 0.94) & \multicolumn{1}{l|}{} & (2, 1) & (6, 1) & (5, 1) & \\ 
    \cline{1-2}\cline{4-6}
    classic & morning & & & too & dull & \\
    (3, 0.93) & (3, 0.0064) & & & (5, 1) & (5, 0.058) & \\ 
    \cline{1-2}\cline{5-6}
    interesting & try & & & always & slow & \\
    (3, 0.085) & (3, 0.0026) & & & (4, 0.89) & (3, 1) & \\ 
    \cline{1-2}\cline{5-6}
    lovely & job & & & never & boring & \\
    (3, 0.99) & (3, 0.99) & & & (3, 1) & (3, 0.0059) & \\ 
    \cline{1-2}\cline{5-6}
    strong & work & & & sometimes & small & \\
    (2, 1) & (2, 0.99) & & & (2, 1) & (2, 1) & \\ 
    \cline{1-2}\cline{5-6}
    bad & thing & & & quite & dark & \\
    (2, 8e-05) & (2, 0.0041) & & & (2, 0.98) & (2, 1) & \\ 
    \cline{1-2}\cline{5-6}
    fun & plan & & & slightly & expensive & \\
    (2, 0.87) & (2, 0.0017) & & & (2, 1) & (2, 1) & \\ 
    \cline{1-2}\cline{5-6}
    funny & lord & & & damn & \multicolumn{1}{l}{} & \\
    (2, 0.55) & (2, 0.007) & & & (2, 0.018) & \multicolumn{1}{l}{} & \\ 
    \cline{1-2}\cline{5-5}
    excellent & question & & \multicolumn{1}{l}{} & \multicolumn{1}{l}{} & \multicolumn{1}{l}{} & \\
    (2, 0.88) & (2, 0.0012) & & \multicolumn{1}{l}{} & \multicolumn{1}{l}{} & \multicolumn{1}{l}{} & \\ 
    \cline{1-2}
    wonderful & walk & & \multicolumn{1}{l}{} & \multicolumn{1}{l}{} & \multicolumn{1}{l}{} & \\
    (2, 0.93) & (2, 0.22) & & \multicolumn{1}{l}{} & \multicolumn{1}{l}{} & \multicolumn{1}{l}{} & \\ 
    \cline{1-2}
    decent & answer & & \multicolumn{1}{l}{} & \multicolumn{1}{l}{} & \multicolumn{1}{l}{} & \\ 
    (2, 0.48) & (2, 0.011) & & \multicolumn{1}{l}{} & \multicolumn{1}{l}{} & \multicolumn{1}{l}{} & \\ 
    \cline{1-2}
    scary & thoughts & & \multicolumn{1}{l}{} & \multicolumn{1}{l}{} & \multicolumn{1}{l}{} & \\
    (2, 0.001) & (2, 0.07) & & \multicolumn{1}{l}{} & \multicolumn{1}{l}{} & \multicolumn{1}{l}{} & \\ 
    \cline{1-2}
    \multicolumn{1}{l|}{} & advice & & \multicolumn{1}{l}{} & \multicolumn{1}{l}{} & \multicolumn{1}{l}{} & \\
    \multicolumn{1}{l|}{} & (2, 0.0052) & & \multicolumn{1}{l}{} & \multicolumn{1}{l}{} & \multicolumn{1}{l}{} & \\ 
    \cline{2-2}
    \multicolumn{1}{l|}{} & mood & & \multicolumn{1}{l}{} & \multicolumn{1}{l}{} & \multicolumn{1}{l}{} & \\
    \multicolumn{1}{l|}{} & (2, 0.99) & & \multicolumn{1}{l}{} & \multicolumn{1}{l}{} & \multicolumn{1}{l}{} & \\ 
    \cline{2-2}
    \multicolumn{1}{l|}{} & timing & & \multicolumn{1}{l}{} & \multicolumn{1}{l}{} & \multicolumn{1}{l}{} & \\
    \multicolumn{1}{l|}{} & (2, 0.2) & & \multicolumn{1}{l}{} & \multicolumn{1}{l}{} & \multicolumn{1}{l}{} & \\
    \cline{2-2}
  \end{tabular}%
  }
\caption[Resampling examples for SST-2]{Resampled words for the example explanations in Table \ref{tab:example_explanations}.
The header of a column indicates which word was replaced.
The word entry in the row shows the replacement.
The numbers in brackets are how often this word was the replacement out of 100 samples (weight) and the prediction of the positive sentiment neuron, which is the true label.
Only words which were sampled at least twice are presented, the columns are ordered by sampling count.}
\label{tab:sampling_examples}
\end{table*}
\mbox{}
}

\newcommand{\insertmnlicorrelationtable}{
    \begin{table*}[tb!]
    \centering
        \begin{tabular}{@{}l|cc|cc|ccc@{}}
        \toprule
        {} & \multicolumn{4}{c|}{Occlusion} & \multicolumn{3}{c}{Gradient} \\
         Method & \textbf{OLM} & \textbf{OLM-S} & Del & UNK & Sen & G*I & IG \\
         \midrule
         \textbf{OLM} & 1.00 & 0.61 & 0.60 & 0.58 & 0.27 & -0.03 & 0.28 \\
         \textbf{OLM-S} & 0.61 & 1.00 & 0.32 & 0.32 & 0.35 & -0.01 & 0.20 \\
         \midrule
         Delete & 0.60 & 0.32 & 1.00 & 0.73 & 0.23 & -0.05 & 0.34 \\
         UNK & 0.58 & 0.32 & 0.73 & 1.00 & 0.22 & -0.03 & 0.32 \\
         \midrule
         Sensitivity Analysis & 0.27 & 0.35 & 0.23 & 0.22 & 1.00 & 0.03 & 0.17 \\
         Gradient*Input & -0.03 & -0.01 & -0.05 & -0.03 & 0.03 & 1.00 & 0.00 \\
         Integrated Gradients & 0.28 & 0.20 & 0.34 & 0.32 & 0.17 & 0.00 & 1.00 \\
        \bottomrule
        \end{tabular}
    \caption[Correlation of explanation methods on MNLI]{Correlation between explanation methods on MNLI development set.
    The table is symmetrical.
    The first two rows are our own methods.
    The next two rows are other occlusion methods.
    The last three rows are gradient-based explanation methods.
    The correlation of different methods is highest between the occlusion methods but never close to 1.
    Gradient*Input does not correlate with any method.}
    \label{tab:mnli_method_correlation}
    \end{table*}
}

\newcommand{\insertsstcorrelationtable}{
    \begin{table*}[tb!]
    \centering
        \begin{tabular}{@{}l|cc|cc|ccc@{}}
        \toprule
         {} & \multicolumn{4}{c|}{Occlusion} & \multicolumn{3}{c}{Gradient} \\
         Method & \textbf{OLM} & \textbf{OLM-S} & Del & UNK & Sen & G*I & IG \\
         \midrule
         \textbf{OLM} & 1.00 & 0.78 & \textbf{0.52} & \textbf{0.47} & 0.30 & 0.02 & 0.35 \\
         \textbf{OLM-S} & 0.78 & 1.00 & \textbf{0.39} & \textbf{0.38} & 0.37 & 0.01 & 0.30 \\
         \midrule
         Delete & \textbf{0.52} & \textbf{0.39} & 1.00 & 0.64 & 0.21 & 0.01 & 0.37 \\
         UNK & \textbf{0.47} & \textbf{0.38} & 0.64 & 1.00 & 0.18 & 0.03 & 0.36 \\
         \midrule
         Sensitivity Analysis & 0.30 & 0.37 & 0.21 & 0.18 & 1.00 & 0.03 & 0.13 \\
         Gradient*Input & 0.02 & 0.01 & 0.01 & 0.03 & 0.03 & 1.00 & 0.04 \\
         Integrated Gradients & 0.35 & 0.30 & 0.37 & 0.36 & 0.13 & 0.04 & 1.00 \\
        \bottomrule
        \end{tabular}
    \caption[Correlation of explanation methods on SST-2]
    {Correlation between explanation methods on SST-2 development set.
    The results resemble the results from Table \ref{tab:mnli_method_correlation}.
    The correlation between \emph{OLM} and other occlusion methods is a little lower.
    In contrast, the correlation between \emph{OLM-S} and other occlusion methods is a little higher.}
    \label{tab:sst2_method_correlation}
    \end{table*}
}

\newcommand{\insertcolacorrelationtable}{
    \begin{table*}[b!]\centering
        \begin{tabular}{@{}l|cc|cc|ccc@{}}
        \toprule
         {} & \multicolumn{4}{c|}{Occlusion} & \multicolumn{3}{c}{Gradient} \\
         Method & \textbf{OLM} & \textbf{OLM-S} & Del & UNK & Sen & G*I & IG \\
         \midrule
         \textbf{OLM} & 1.00 & 0.56 & \textbf{0.25} & \textbf{0.21} & 0.20 & 0.02 & 0.15 \\
         \textbf{OLM-S} & 0.56 & 1.00 & \textbf{0.15} & \textbf{0.12} & 0.29 & 0.03 & 0.09 \\
         \midrule
         Delete & \textbf{0.25} & \textbf{0.15} & 1.00 & 0.35 & 0.02 & 0.04 & 0.18 \\
         Unk & \textbf{0.21} & \textbf{0.12} & 0.35 & 1.00 & 0.03 & 0.03 & 0.14 \\
         \midrule
         Sensitivity Analysis & 0.20 & 0.29 & 0.02 & 0.03 & 1.00 & -0.00 & 0.07 \\
         Gradient*Input & 0.02 & 0.03 & 0.04 & 0.03 & -0.00 & 1.00 & 0.12 \\
         Integrated Gradients & 0.15 & 0.09 & 0.18 & 0.14 & 0.07 & 0.12 & 1.00 \\
        \bottomrule
        \end{tabular}
    \caption[Correlation of explanation methods on CoLA]
    {Correlation between explanation methods on CoLA development set.
    The correlation between \emph{OLM} and \emph{OLM-S} and other occlusion methods (in bold) is much lower than on the other two tasks.
    This indicates that this dataset could be a corner case for our method.
    Correlation between other methods is also lower but to a smaller extend.}
    \label{tab:cola_method_correlation}
    \end{table*}
}

\newcommand{\insertcolasignificancetable}{
    \begin{table*}[tb!]\centering
        \begin{tabular}{@{}l|lll@{}}
        \toprule
         {} & \multicolumn{3}{c}{relevance aggregation} \\
         {} & Avg. & Sum & Max \\
         \midrule
         unacceptable sentence & 0.275 & 1.89 & 0.893 \\
         acceptable sentence & 0.0384 & 0.304 & 0.172 \\
         \midrule
         p-value & <0.001 & <0.001 & <0.001 \\
         \bottomrule
        \end{tabular}
    \caption[Significance test for explanations on CoLA]
    {Relevance values accumulated over inputs by the method in the column header, averaged over all sentences with correct classification and probability $p \geq 0.9$.
    For all accumulations the averages differ dramatically, which indicates that the resamples changed the prediction a lot more for unacceptable sentences.
    A \emph{Welch's t-test} was performed to compare the means and yielded a $p\text{-value}<0.001$ for all three methods.}
    \label{tab:cola_significance}
    \end{table*}
}

\newcommand{\insertcolaexampletable}{
\begin{table*}[b!]
  \resizebox{\columnwidth}{!}{
  \centering
  \begin{tabular}{l|l|l}
    id&relevances&max value \\ \midrule
    1&\colorbox[RGB]{255,254,254}{\strut John} \colorbox[RGB]{255,17,17}{\strut paid} \colorbox[RGB]{255,58,58}{\strut me} \colorbox[RGB]{255,0,0}{\strut against} \colorbox[RGB]{255,254,254}{\strut the} \colorbox[RGB]{255,180,180}{\strut book} \colorbox[RGB]{254,254,255}{\strut .} &0.99\\
    2&\colorbox[RGB]{255,254,254}{\strut The} \colorbox[RGB]{255,254,254}{\strut person} \colorbox[RGB]{255,0,0}{\strut confessed} \colorbox[RGB]{255,33,33}{\strut responsible} \colorbox[RGB]{255,254,254}{\strut .} &1\\
    3&\colorbox[RGB]{255,254,254}{\strut Medea} \colorbox[RGB]{255,0,0}{\strut tried} \colorbox[RGB]{255,254,254}{\strut the} \colorbox[RGB]{255,177,177}{\strut nurse} \colorbox[RGB]{255,254,254}{\strut to} \colorbox[RGB]{254,254,255}{\strut poison} \colorbox[RGB]{254,254,255}{\strut her} \colorbox[RGB]{255,254,254}{\strut children} \colorbox[RGB]{255,254,254}{\strut .} &0.92\\
    \midrule
    4&\colorbox[RGB]{255,229,229}{\strut to} \colorbox[RGB]{255,0,0}{\strut die} \colorbox[RGB]{255,254,254}{\strut is} \colorbox[RGB]{255,236,236}{\strut no} \colorbox[RGB]{255,180,180}{\strut fun} \colorbox[RGB]{254,254,255}{\strut .} &0.49\\
    5&\colorbox[RGB]{255,254,254}{\strut This} \colorbox[RGB]{255,0,0}{\strut teacher} \colorbox[RGB]{254,254,255}{\strut is} \colorbox[RGB]{255,254,254}{\strut a} \colorbox[RGB]{255,205,205}{\strut genius} \colorbox[RGB]{255,254,254}{\strut .} &0.056\\
    6&\colorbox[RGB]{255,116,116}{\strut Soaring} \colorbox[RGB]{255,160,160}{\strut temperatures} \colorbox[RGB]{255,254,254}{\strut are} \colorbox[RGB]{255,0,0}{\strut predicted} \colorbox[RGB]{255,254,254}{\strut for} \colorbox[RGB]{254,254,255}{\strut this} \colorbox[RGB]{255,254,254}{\strut weekend} \colorbox[RGB]{255,254,254}{\strut .} &0.08\\
  \end{tabular}
  }
  \caption[Example explanations for CoLA]
  {Six randomly selected example sentences from the CoLA dataset with explanations.
  The upper three examples are grammatically unacceptable, the lower three examples are acceptable.
  All examples are correctly classified with a probability of at least $0.9$.
  The upper examples have words with much higher relevance as the resampled words can make the sentence acceptable.
  In the lower examples the resampled words change the prediction only once.
  Sentences with likely replacement words that changed the prediction can be found in Table \ref{tab:resamples_cola}.
  }
  \label{tab:example_explanations_cola}
\end{table*}
}

\newcommand{\insertcolaresamples}{
\begin{table*}[tb!]
  \resizebox{\columnwidth}{!}{
  \centering
  \begin{tabular}{l|l|l|l}
    \toprule
    id & orig.~word & new sentence & prediction \\ 
    \midrule
    1 & paid & John \emph{pushed} me against the book . & 4e-4 \\
    1 & paid & John \emph{pressed} me against the book . & 3.6e-4 \\
    1 & me & John paid \emph{damages} against the book . & 4.3e-4 \\
    1 & me & John paid \emph{taxes} against the book . & 7.9e-4 \\
    1 & against & John paid me \emph{for} the book . & 3.7e-4 \\
    2 & confessed & The person \emph{was} responsible . & 4.4e-4 \\
    2 & confessed & The person \emph{is} responsible . & 4.4e-4 \\
    2 & responsible & The person confessed \emph{himself} . & 4e-4 \\
    3 & tried & Medea \emph{orders} the nurse to poison her children . & 3.5e-4 \\
    3 & nurse & Medea tried the \emph{same} to poison her children . & 1.1e-3 \\
    \midrule
    4 & die & to \emph{eat} is no fun . & 0.089 \\
    \bottomrule
  \end{tabular}
  }
  \caption[Resampling examples for CoLA]
  {Resampled sentences for Table \ref{tab:example_explanations_cola}.
  All resampled words that changed the predicted class and were sampled at least 10 out of 100 times are depicted.
  Most of the previously unacceptable sentences are now acceptable and thus correctly classified by the model.
  In the last row we can see that the only frequent sample for acceptable sentences that changed the prediction was misclassified by the model.
  The upper rows of the table show that the interval of predictions for acceptable samples is very narrow.
  They have a probability between $0.035\%$ and $0.11\%$ of being unacceptable, which is only a difference by factor three.
  }
  \label{tab:resamples_cola}
\end{table*}
}

\newcommand{\insertfavasignificancetable}{
    \begin{table*}[tb!]
    \centering
        \begin{tabular}{@{}l|ll@{}}
        \toprule
         {} & unacceptable & acceptable \\
         \midrule
         average verb relevance & 0.400 & 0.0960 \\
         average word relevance & 0.115 & 0.0429 \\
         \midrule
         p-value & <0.001 & <0.001  \\
         \bottomrule
        \end{tabular}
    \caption[Significance test for explanations on FAVA]
    {Relevance values averaged over all sentences with correct classification and probability $p \geq 0.9$.
    For both unacceptable and acceptable sentences the verb is significantly more relevant than the average word.
    A \emph{Welch's t-test} was performed to compare the means and yielded a $p\text{-value}<0.001$ for both averages.}
    \label{tab:fava_significance}
    \end{table*}
}

\newcommand{\insertfavaexampletable}{
\begin{table*}[tb!]
  \resizebox{\columnwidth}{!}{
  \centering
  \begin{tabular}{l|l|l}
    verb frame & relevances & max value \\ \midrule
    causative- & \colorbox[RGB]{255,136,136}{\strut the} \colorbox[RGB]{255,143,143}{\strut chapter} \colorbox[RGB]{255,0,0}{\strut edited} \colorbox[RGB]{255,252,252}{\strut .} &0.49\\
    inchovative & \colorbox[RGB]{255,0,0}{\strut david} \colorbox[RGB]{255,169,169}{\strut edited} \colorbox[RGB]{255,245,245}{\strut the} \colorbox[RGB]{255,220,220}{\strut chapter}
    \colorbox[RGB]{255,254,254}{\strut .} &0.18\\
    \midrule
    spray- & \colorbox[RGB]{253,253,255}{\strut michael} \colorbox[RGB]{255,0,0}{\strut poured} \colorbox[RGB]{255,227,227}{\strut the} \colorbox[RGB]{255,184,184}{\strut bucket} \colorbox[RGB]{255,21,21}{\strut with} \colorbox[RGB]{255,252,252}{\strut the} \colorbox[RGB]{255,233,233}{\strut soup} \colorbox[RGB]{255,255,255}{\strut .} &0.85\\
    load & \colorbox[RGB]{255,0,0}{\strut michael} \colorbox[RGB]{253,253,255}{\strut poured} \colorbox[RGB]{255,244,244}{\strut the} \colorbox[RGB]{255,161,161}{\strut soup} \colorbox[RGB]{255,58,58}{\strut into} \colorbox[RGB]{255,250,250}{\strut the} \colorbox[RGB]{252,252,255}{\strut bucket} \colorbox[RGB]{254,254,255}{\strut .} &0.028\\
    \midrule
    there- & \colorbox[RGB]{255,154,154}{\strut there} \colorbox[RGB]{255,65,65}{\strut agreed} \colorbox[RGB]{255,241,241}{\strut with} \colorbox[RGB]{255,242,242}{\strut the} \colorbox[RGB]{255,253,253}{\strut politician} \colorbox[RGB]{255,220,220}{\strut a} \colorbox[RGB]{255,0,0}{\strut protester} \colorbox[RGB]{255,254,254}{\strut .} &0.082\\
    load & \colorbox[RGB]{246,246,255}{\strut a} \colorbox[RGB]{255,0,0}{\strut protester} \colorbox[RGB]{255,174,174}{\strut agreed} \colorbox[RGB]{255,233,233}{\strut with} \colorbox[RGB]{255,245,245}{\strut the} \colorbox[RGB]{255,162,162}{\strut politician} \colorbox[RGB]{255,254,254}{\strut .} &0.15\\
    \midrule
    understood- & \colorbox[RGB]{255,54,54}{\strut kelly} \colorbox[RGB]{255,0,0}{\strut joked} \colorbox[RGB]{255,68,68}{\strut david} \colorbox[RGB]{254,254,255}{\strut .} &0.84\\
    object & \colorbox[RGB]{255,236,236}{\strut kelly} \colorbox[RGB]{255,255,255}{\strut and} \colorbox[RGB]{255,218,218}{\strut david} \colorbox[RGB]{255,0,0}{\strut joked} \colorbox[RGB]{254,254,255}{\strut .} &0.42\\
    \midrule
    dative & \colorbox[RGB]{255,225,225}{\strut nicole} \colorbox[RGB]{255,163,163}{\strut proclaimed} \colorbox[RGB]{255,232,232}{\strut the} \colorbox[RGB]{255,253,253}{\strut greatest} \colorbox[RGB]{255,107,107}{\strut athlete} \colorbox[RGB]{255,0,0}{\strut to} \colorbox[RGB]{255,200,200}{\strut rebecca} \colorbox[RGB]{255,254,254}{\strut .} &0.43\\
    & \colorbox[RGB]{255,136,136}{\strut nicole} \colorbox[RGB]{255,0,0}{\strut proclaimed} \colorbox[RGB]{255,111,111}{\strut rebecca} \colorbox[RGB]{255,254,254}{\strut the} \colorbox[RGB]{255,186,186}{\strut greatest} \colorbox[RGB]{255,135,135}{\strut athlete} \colorbox[RGB]{255,254,254}{\strut .} &0.25\\
  \end{tabular}
  }
  \caption[Example explanations for FAVA]
  {Randomly selected example explanations for sentence pairs in the FAVA dataset.
  The examples are ordered by verb frame, the unacceptable example is always the first.
  All examples are correctly classified with probability $p \geq 0.9$.
  The verbs are frequently identified as the most important word in a sentence.}
  \label{tab:example_explanations_fava}
\end{table*}
}

\newcommand{\insertfavacorrelationfigure}{
\begin{figure}[tb!]
  \centering
  \includegraphics[trim=0 10 0 10, scale=0.9]{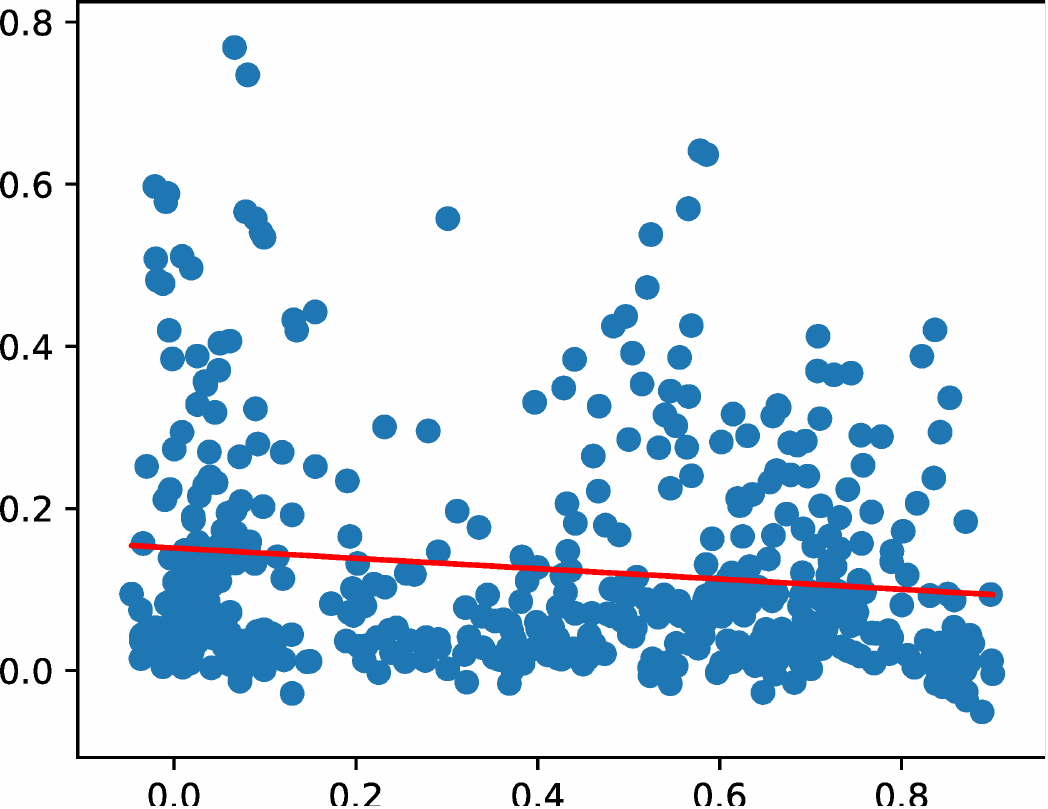}
  \caption[Correlation of verb relevance for FAVA sentence pairs]
  {Correlation of verb relevance for FAVA sentence pairs.
  The x-axis denotes the relevance of the verb in the unacceptable sentence of a sentence pair.
  The y-axis denotes the relevance of the verb in the acceptable sentence of a sentence pair.
  Only pairs are depicted where both sentences were correctly predicted with probability $p \geq 0.9$.
  Thus, the interval of possible values for both axes is $[-0.1, 1]$.}
  \label{fig:verb_correlation}
\end{figure}
}

\frontmatter

\pagestyle{plain}

\begin{titlepage}
\begin{center}

\vspace*{.06\textheight}
{\scshape\Large \univname\par}\vspace{1.5cm}
\textsc{\large Master's Thesis}\\[0.5cm]

\HRule \\[0.4cm]
{\LARGE \bfseries \ttitle\par}\vspace{0.4cm}
\HRule \\[1.5cm]
 
\begin{minipage}[t]{0.45\textwidth}
\begin{flushleft} \normalsize
\emph{Author:}\\
\authorname
\end{flushleft}
\end{minipage}
\begin{minipage}[t]{0.45\textwidth}
\begin{flushright}
\emph{Supervisors:} \\
\supname
\end{flushright}
\end{minipage}\\[1.5cm]
 
\vfill

\textit{A thesis submitted in fulfillment of the requirements\\ for the degree of \degreename\thinspace in}\\[0.3cm]
\progname\\[0.3cm]
\textit{at the}\\[0.3cm]
\facname\\[1.5cm]
 
\vfill

{September 27, 2020}\\
 
\end{center}
\end{titlepage}

\pdfbookmark[0]{Preliminaries}{Preliminaries}
\begin{declaration}
\addsectiontocentry{\authorshipname}
\vspace{2cm}

\noindent I, \authorname, declare that this thesis, titled \enquote{\thesis}, and the work presented in it are my own.
I confirm that I worked independently and only with the sources and aids indicated.\footnote{Nevertheless, this thesis is written in first-person plural.}
All passages of the work which I have taken from these sources and aids, either in wording or in meaning, are marked and listed in the bibliography.
Parts of the presented research have been published with a co-author \citep{harbecke2020considering}.\footnote{I hereby acknowledge the Association for Computational Linguistics for letting me describe findings of publications co-authored by me.}
Content contributions of the co-author that are mentioned are marked as such.
I am familiar with \enquote{Richtlinie zur Sicherung guter wissenschaftlicher Praxis für Studierende an der Universität Potsdam (Plagiatsrichtlinie)}\footnote{\url{https://www.uni-potsdam.de/am-up/2011/ambek-2011-01-037-039.pdf}}.\\[2cm]
 
\noindent Date:\\
\rule[0.5em]{25em}{0.5pt}

\noindent Signed:\\
\rule[0.5em]{25em}{0.5pt}
\end{declaration}

\begin{abstractpage}
\renewcommand{\abstractname}{English Abstract}
\begin{abstract}{}
    \addsectiontocentry{\abstractname}\\
    Deep neural networks are powerful statistical learners.
    However, their predictions do not come with an explanation of their process.
    To analyze these models, explanation methods are being developed.
    We present a novel explanation method, called \emph{OLM}, for natural language processing classifiers.
    This method combines occlusion and language modeling, which are techniques central to explainability and NLP, respectively.
    \emph{OLM} gives explanations that are theoretically sound and easy to understand.
    
    We make several contributions to the theory of explanation methods.
    Axioms for explanation methods are an interesting theoretical concept to explore their basics and deduce methods.
    We introduce a new axiom, give its intuition and show it contradicts another existing axiom.
    Additionally, we point out theoretical difficulties of existing gradient-based and some occlusion-based explanation methods in natural language processing.
    We provide an extensive argument why evaluation of explanation methods is difficult.
    We compare \emph{OLM} to other explanation methods and underline its uniqueness experimentally.
    Finally, we investigate corner cases of \emph{OLM} and discuss its validity and possible improvements.

\end{abstract}

\renewcommand{\abstractname}{Deutsche Zusammenfassung}
\begin{abstract}{}
    \addsectiontocentry{\abstractname}\\
    Tiefe neuronale Modelle sind gut im statistischen Lernen.
    Jedoch liefern deren Vorhersagen keine Erklärungen des Vorgangs.
    Um diese Modelle zu analysieren, hat man Erklärungstechniken entwickelt.
    Wir präsentieren eine neue Erklärungstechnik, gennant \emph{OLM}, für klassifizerende Modelle linguistischer Datenverarbeitung.
    Diese Methode kombiniert das Maskieren von Merkmalen mit Sprachmodellierung, was jeweils grundlegende Methoden der Erklärungstechnik und linguistischer Datenverarbeitung sind.
    \emph{OLM} liefert Erklärungen die theoretisch fundiert und einfach zu verstehen sind.
    
    Wir machen mehrere theoretische Beiträge zu Erklärungstechniken.
    Axiome für Erklärungstechniken sind ein interessantes theoretisches Konzept um deren Grundlagen auszuloten und Techniken abzuleiten.
    Wir führen ein neues Axiom ein, legen die Intuition dar und zeigen dass es einem existierenden Axiom widerspricht.
    Zusätzlich zeigen wir theoretische Problematik in der Anwendung von gradienten- und manchen maskierungsbasierten Erklärungstechniken bei linguistischer Datenverarbeitung auf.
    Wir argumentieren umfangreich, warum die Evaluation von Erklärungstechniken schwierig ist.
    Wir vergleichen \emph{OLM} mit anderen Erklärungstechniken und heben dessen Alleinstellungsmerkmal hervor.
    Abschließend betrachten wir Grenzfälle der Anwendung von \emph{OLM} und diskutieren dessen Validität und mögliche Verbesserungen.

\end{abstract}
\end{abstractpage}

\tableofcontents
\listoffigures
\begingroup
\let\clearpage\relax
\listoftables
\endgroup

\begin{symbols}{ll}

$p_{data}$ & probability distribution of data \\
$p_{LM}$ & probability distribution of a language model \\

\addlinespace

$f_{\theta}$ & neural network with parameters $\theta$ \\
$f$ & neural network, or, in general, black-box function \\
$f_c:=\text{proj}_c \circ f$ & projection of function $f$ to class $c$, i.e.~output neuron $c$ \\

\addlinespace

$(X, Y)$ & labeled dataset \\
$X$ & a dataset or, more general, the whole input space \\
$Y$ & label set \\
$x \in X$ & an input element of the dataset; an input vector  \\
$x_i$ & an indexed feature of the input $x$ \\
$x_{\setminus i}$ & an input without the feature at $i$-th position \\
$(x_{\setminus i}, \hat{x}_i)$ & an input with a replacement feature \\

\addlinespace

$r_{f, c}(x_i)$ & relevance of an input feature $x_i$ regarding function $f$ \\ 
& \quad and class $c$\\
\end{symbols}

\mainmatter

\pagestyle{thesis}

\chapter{Introduction}

\label{ch:Introduction}

The advent of deep learning has created an explanation gap as the models are considered hard to interpret \citep{guidotti2018survey, adadi2018peeking}.
Models without hand-engineered features are hard to understand in their decision making.
This makes explanation methods highly relevant.
In natural language processing most state-of-the-art architectures are neural networks.
To understand their decisions we point out gaps in existing explanation methods and develop a novel method.

\section{Deep Learning}

Deep learning describes the architectures of multi-layered neural networks and methods to train them.
Deep neural networks (DNNs) learn features from data.
The layers of a deep neural network learn increasingly higher level features \citep{deng2014deep}.

To introduce deep neural networks we first motivate interest in them.
Then, we introduce some of their theory by explaining a possible deep neural network architecture and how DNNs can be trained.
Lastly, we discuss specific properties that show DNNs' relevance to the presented work.

\subsection{Motivation}

Deep neural networks have achieved state-of-the-art performance on a wide variety of tasks, such as
\begin{itemize}
    \item image recognition \citep{ciregan2012multi, krizhevsky2012imagenet},
    \item text classification \citep{kim2014convolutional},
    \item perfect information games \citep{silver2016mastering, silver2018general},
    \item or estimating the underlying probability distribution of a sample space \citep{bengio2003neural, goodfellow2014generative}.
\end{itemize}
In addition to the wide variety of tasks they can perform well, the prediction process of a neural network can be seen as opaque.
A neural network learns its parameters from data and does not need to have feature extractions engineered by humans.
Thus, external techniques are required to explain the training and decisions of a neural network.

\insertneuralnetwork

\subsection{Architecture}
A neural network consists of neurons in layers.
Figure \ref{fig:neural_network} gives a schematic view of a neural network.
The input layer displays an input vector $x=(x_1, x_2, x_3)^T$ with three input dimensions.
For the first hidden layer this input vector is multiplied by a weight matrix $W^1$ with $3\times 4$ dimensions, a bias vector $b^1=(b^1_1, b^1_2, b^1_3, b^1_4)^T$ is added and a non-linear activation function $\sigma^1$ is applied component-wise.
This is repeated for the next hidden layer and the output layer with different weights and biases.
Therefore, a step from a layer to the next is an affine transformation followed by an activation function.
The activation function of the output layer is usually chosen such that the output neurons represent a probability distribution or individual probability functions, depending on the formulation of the problem and the data.
A mathematical formulation of the network would be
\begin{equation}
    \hat{y} = \sigma^3(W^3 \sigma^2(W^2 \sigma^1(W^1 x + b^1) + b^2) + b^3).
\end{equation}

A neural network with several hidden layers is called deep neural network (DNN).
The weights and biases are the trainable parameters $\theta$ of a neural network.
The prediction $\hat{y}$ depends on these parameters, also referred to as weights.
A simpler formulation if we are not interested in specific weights is
\begin{equation}
    \hat{y} = f_{\theta}(x).
\end{equation}

\subsection{Training}
\label{sec:training}
To train a DNN we need an objective and a loss function.
The objective is usually a labeled dataset $(X, Y)$ and tells us what the network should predict for each input.
An element of the dataset $x\in X$ is usually a coordinate vector over $\mathbb{R}$.
The loss function is a function $L(\hat{y}, y)$ of the prediction $\hat{y}$ and true label $y\in Y$.
Sometimes a regularizer, which is a loss function on the network parameters $\theta$, also usually coordinate vectors over $\mathbb{R}$, is added to this loss function.

With all these ingredients, a neural network is usually trained with backpropagation \citep{linnainmaa1970representation, rumelhart1986learninga, rumelhart1985learningb} and a variation of gradient descent \citep{cauchy1847methode}.
Backpropagation is a method that propagates the error $E$ calculated by the loss function to the parameters $\theta$ of the DNN via the chain rule of differentiation.
The need for differentiability explains why often both the inputs and parameters are coordinate vectors over $\mathbb{R}$.
It ensures that the gradients are also real valued.
This is important for section \fullref{sec:nlp_repr}.

Differentiation can be done in parallel for all parameters of a layer.
Gradient descent is an optimization technique of these parameters.
Let us view the neural network as a high-dimensional function over all parameters $\theta$.
It changes the parameter values by stepping proportionally to the size of the partial derivative of the loss over the whole dataset regarding the parameter in each direction.
Gradient descent can be seen an optimization alternative to using Newton's Method \citep{newton1736method, raphson1690analysis, simpson1740essays}, which looks for zeros of a function, for the first derivative of a function.

Gradient descent only updates the parameters once per iteration over the dataset.
This is inefficient as subsets of the dataset (mini-batch) can provide a good estimation of the gradient \citep{wilson2003general, bottou2008tradeoffs}.
Stochastic gradient descent \citep{robbins1951stochastic} averages the gradients over a mini-batch and does one optimization step for this mini-batch.
This is a better trade-off between update time and update quality.
There are many popular and recent variants and alternatives to stochastic gradient descent such as Momentum \citep{qian1999momentum} and ADAM \citep{kingma2014adam}.

\subsection{Capabilities}
\label{sec:capabilites}
The universal approximation theorem states that with specific restrictions for the width \citep{lu2017expressive} or depth \citep{hanin2017universal} DNNs can approximate any continuous convex function.
This gives an intuition on why they are state-of-the-art for many prediction tasks.
Furthermore, \cite{choromanska2015loss} show that deep networks have better loss surfaces than shallow networks for training.
This means that the local optima that optimizers find are closer to the global optimum for deep networks.
This is underlined by the information bottleneck principle \citep{tishby2015deep, shwartz2017opening} which states that training of a DNN is faster than that of similarly capable shallow networks.
The success of deep learning is also partly due to hardware with parallel computing capabilities \citep{strigl2010performance}.

For this work the relevance of DNNs is two-fold.
First, we try to explain their behaviour when performing state-of-the-art prediction on natural language processing tasks.
Second, we use neural language models to create these explanations.

\section{Natural Language Processing}
\label{sec:nlp}
Natural language processing (NLP) encompasses the intersection between human language and the processing of it by computing machinery.
The field of NLP is nowadays mostly concerned with a statistical and quantitative processing and modeling of mass amounts of language data.
\cite{manning1999foundations} state:
\blockquote{\enquote{Increasingly, businesses, government agencies and individuals are confronted with large amounts of text that are critical for working and living, but not well enough understood to get the enormous value out of them that they potentially hide.}}
This development has been amplified by the success of DNNs which are currently state of the art for many of the popular datasets of these tasks.
This includes
\begin{itemize}
    \item automatic speech recognition on the LibriSpeech corpus \citep{panayotov2015librispeech} by \cite{synnaeve2019end} and the Wall Street Journal corpus \citep{paul1992design} by \cite{povey2016purely},
    \item part-of-speech tagging on the Wall Street Journal part of the Penn Treebank \citep{mitchell1999treebank} by \cite{bohnet2018morphosyntactic},
    \item sentiment analysis on the IMDB dataset \citep{maas2011learning}, the Stanford Sentiment Treebank \citep{socher2013recursive} and the Yelp Review dataset \citep{zhang2015character} by \textsc{XLNet} \citep{yang2019xlnet},
    \item text classification on the AG News corpus, the DBpedia onthology \citep{zhang2015character} and the TREC dataset \citep{voorhees2000building} by \textsc{XLNet}  \citep{yang2019xlnet}.
    \item There are also multi-task benchmarks, such as GLUE \citep{wang2019glue} and SuperGLUE \citep{wang2019superglue}, that combine several NLP classification tasks.
    Leading both benchmarks is \textsc{T5} \citep{raffel2019exploring}.
\end{itemize}

At least three things are notable regarding the previous list.
First, the variety of tasks is wide, ranging from speech recognition on audio data and generating text data in language modeling to classifying words, sentences or texts in part-of-speech tagging, relationship extraction, sentiment analysis and text classification.
This makes the dominance of neural architectures in these tasks even more impressive.

Second, \emph{XLNet} \citep{yang2019xlnet} and \emph{T5} \citep{raffel2019exploring} appear frequently on top of the leaderboard on many of these tasks.
Both use a pre-trained language model, i.e.~they learn a representation of language by going over large text datasets with billions of words, such as the BooksCorpus \citep{zhu2015aligning} or English Wikipedia\footnote{\url{https://en.wikipedia.org}}.
The training of language models will be described in more detail in section \fullref{sec:Language_Modeling}.
\cite{raffel2019exploring} even encode the problem formulation into the representation.

Third, most of the datasets are classification tasks in some sense where the neural model has a preselected set of outputs.
These are the problems and models we are interested in.
The method presented in this thesis yields explanations for NLP classification tasks.

\subsection{Language Representations for Deep Learning}
\label{sec:nlp_repr}
\epigraph{Time flies like an arrow;\\
fruit flies like bananas}{Anthony Oettinger}

An important component to utilize neural networks in NLP is the representation of the input.
This is nontrivial, as we saw in section \fullref{sec:training}.
We have to assign real valued coordinate vectors to our input.
Furthermore, the unit of language, called token or atomic parse element, from which to map into our vector space is nonobvious.
It is imperative to choose a token that allows for an unambiguous mapping.
For written language characters, words, sentences and documents are among the candidates for this atomic parse element.

A simple realization of this mapping is to count the words of an input and create a vector representation with vector indices corresponding to words.
This method is called \emph{bag-of-words} \citep{harris1954distributional}.
It can be used both on a sentence and a document level.
A practical improvement on this is what is now called \emph{tf-idf} \citep{salton1975vector}.
There, the word counts are scaled with the logarithm of the inverse of the ratio of documents containing the word.
Both methods do not preserve the order or contextual meaning of words.
Locally, the order and contextuality can be incorporated by choosing to use n-grams instead of or in addition to single words.
However, this does not retain information about longer contexts.
It also yields exponentially more possible n-grams and fewer counts of a specific n-gram for increasing n, which makes the representations less precise and efficient.

A variation to \emph{bag-of-words} is \emph{one-hot-encoding} where the words are not counted but words get assigned pairwise distinct standard basis vectors.
If we ignore out-of-vocabulary words, this is an one-to-one function between inputs and the representation.
This method can be used as underlying transformation for representing words with other methods.
Compared to the previous methods, it allows the order of words in the whole input to be kept.
However, none of these methods provide information about the similarity of words.

\cite{mikolov2013efficient} introduce \emph{word2vec}, a learned vector space representation of words.
It can be called an embedding because the dimensionality of this vector space is lower than the number of words that are represented.
This means that the vectors corresponding to words are not pairwise orthogonal, generally.
Words with similar meaning are supposed to have a small angle between their vector representations.
Words are predicted from their context with the intuition that words that appear in similar contexts have similar meaning.
At the time they achieved state of the art on semantic word similarity.
\emph{GloVe} \citep{pennington2014glove} explicitly learns vector representations that are based on co-occurrence.
These representations allow the DNNs that employ them to use the order of words if they desire.
Thus, context is preserved but does not influence representation itself.

Many approaches use characters \citep{wieting2016charagram, bojanowski2017enriching} or sub-words \citep{wu2016google, kudo2018sentencepiece} as tokens to improve on word representations.
By choosing a smaller unit, similarities between words with similar spelling or the same root can be incorporated into the representation.
Still, it does not enable different representations of the same word in different contexts.

\cite{howard2018universal} use the language modeling described in section \fullref{sec:Language_Modeling} to create a vector embedding where the information of context is fed through several layers.
This enables a different representation for \enquote{flies like} in the epigraph of this section, depending on the context.

\chapter{Explainability of DNNs}

\label{ch:Explainability}

In section \fullref{sec:capabilites} we indicated the performance abilities of DNNs.
However, their state-of-the-art prediction performance is not the measure of all things.
Recital 71 of the European Union's \textit{General Data Protection Regulation}\footnote{\url{https://eur-lex.europa.eu/legal-content/EN/TXT/PDF/?uri=CELEX:32016R0679}} states:
\blockquote{\enquote{[decision-making based solely on automated processing] should be subject to suitable safeguards, which should include specific information to the data subject and the right [\ldots] to obtain an explanation of the decision reached after such assessment and to challenge the decision.}}
Its potential legal ramifications are discussed in \cite{goodman2017european}.
This regulation clearly states that for real-world applications of DNNs explanations are necessary.
Explanations as support for a decision by a black box is detailed in \cite{lombrozo2006structure}: \enquote{explanations are [\ldots] the currency in which we exchanged beliefs}.
We will be using the term explainability (of DNNs) to refer to this field, not the more frequently used \enquote{Explainable Artificial Intelligence (XAI)} or \enquote{Interpretable Machine Learning}.
Both these terms indicate that there is something inherently explainable or interpretable about the models in deep learning.
This may be the case in machine learning but cannot be assumed in general.
Furthermore, it is very controversial whether the \enquote{Artificial Intelligence} moniker is a precise or helpful representation of the characteristics of deep learning \citep{jordan2019artificial}.

Several popular surveys of explainability exist.
We summarize some of them to pigeonhole our work more precisely.
\cite{doshi2017towards} point out that explanations fill an incompleteness in the problem that deep learning models work on.
E.g., the objective given to the model during training may not have measured generalization performance adequately.
This can be uncovered with explanations.
\cite{doshi2017towards} divide explanations into local and global explanations.
Local explanations try to explain the model at a specific input, whereas global explanations try to explain the model as a whole or its general behaviour.

\cite{adadi2018peeking} highlight key concepts in explainability.
They give four reasons for explanations:
\enquote{explain to justify}, \enquote{explain to control}, \enquote{explain to improve} and \enquote{explain to discover}.
Furthermore, they add two categories to sort explanation methods into.
They differentiate between model-specific interpretability and model-agnostic interpretability with the latter denoting methods that can work on any black-box model.
As a subcategory to model-specific methods they discern intrinsic and post-hoc methods.

\cite{guidotti2018survey} present a detailed taxonomy of problems that explanation methods can try to solve.
They give formal definitions of these problems, including intrinsically explainable models and post-hoc black-box methods.
In addition to the previous methods, they describe black-box model explanation, which tries to build intrinsically explained models that approximate the model to be explained.

There is an extended discussion of explanation and psychology in \cite{miller2019explanation}.
They argue that many of the insights gained in psychological understanding of explanations can be used for explainability of DNNs.
We will try to psychologically motivate our method in section \fullref{sec:olm_motivation}.
However, we disagree that many insights from psychology can be easily transferred.
Two main questions for explanations of black boxes are how to generate explanations from them and how to evaluate the faithfulness of their explanations.
This removes assumptions generally held in psychological contexts.

There exist a variety of NLP specific explanations.
\cite{li2016visualizing} present methods to view and test heatmaps for different recurrent networks.
They investigate sentiment analysis by putting forward different investigative input.
\cite{alvarez2017causal} create an explanation graph between structured input and output. 
This is especially useful for sequence to sequence tasks like machine translation.
Many state-of-the-art models use attention \citep{bahdanau2014neural, vaswani2017attention}.
Some papers argue whether attention weights as explanations are permissible.
\cite{jain2019attention} show that attention weights do not necessarily agree with other explanations and can also be distorted while not changing the prediction.
\cite{wiegreffe2019attention} disagree and show that under some circumstances these distortion lead to a significant decrease in performance.
Synthesized, these papers argue that attention weights can be used as explanation if and only if the attention is vital for the model. 
There is also a variety of approaches that determine the quality of models by linguistic analysis \citep{linzen2016assessing, mccoy2019right}.

We provide some examples of methods that that do not belong to our category of explanations and will not be discussed further. 
\cite{kim2018interpretability} analyze neural models by determining which concepts were important to a classification decision.
Intrinsic model-specific explanations can be found in older machine learning approaches, such as decision trees, or in model architectures that provide explanations and predictions in parallel \citep{zhang2018interpretable}.
\cite{ribeiro2016should} explain by learning a local explainable model around the prediction of a neural model.

We will now focus on local post-hoc input space explanations.
They describe how much a feature of one specific input contributed to a specific output (class).
They give a real value called \textit{relevance} for every input feature which can be used to create a saliency map \citep{simonyan2013deep}.
An example can be found in Table \ref{tab:example_explanations}.
The output can be the true label neuron, the predicted neuron or another output of choice.
A positive relevance value indicates that the feature contributed positively to the given output, whereas a negative relevance value indicates that the feature distracted from the output.
Input features can be the input values for the first layer, or a cluster of these.
For NLP, e.g., all input values of a word (and punctuation mark) can be clustered.
If the explanation method gives relevances for every input dimension the relevances are aggregated such that every word receives one relevance value.
This is often achieved by summing the relevances \citep{arras2017relevant}.

An important part of explainability is occlusion.
It is either presented as an explanation method \citep{robnik2008explaining, zintgraf2017visualizing} where the difference in prediction when removing an input feature is seen as indicator for the importance of this feature .
Alternatively, it is used as evaluation of explanation methods \citep{zeiler2014visualizing, montavon2018methods}, where it is argued that (other) explanation methods should find features whose occlusion change the prediction significantly.

The use of a method both for generating and evaluating explanations points to a false dichotomy between explanation methods and their evaluation, which we discuss extensively in section \fullref{sec:explanation_evaluation}.
In general, most objective evaluation methods allow for a derivation of an explanation method which satisfies this evaluation perfectly.
This has already been done in \cite{kindermans2017learning} and when developing the theory for this thesis, it was originally intended as an evaluation method.
However, we argue the standard for an evaluation method should be higher.
It should not be considered as just an accessory to an explanation method to justify the explanations.

Explanations are a simplification of the model's decision process.
They try to present the process in a human-understandable way.
For complex models this is always an approximation.
To strictly define some ground rules for this approximation, axioms for explanation methods are developed.

\section{Axioms for Explanation Methods}
Axioms are a proposed method to develop and test explanation methods by \cite{sundararajan2017axiomatic} and extensively discussed in \cite{lipton2018mythos}.
Additional axioms are proposed in \cite{bach2015pixel}, \cite{ancona2017towards}, \cite{kindermans2019reliability} and \cite{srinivas2019full}.
The advantage of axiomatic analysis is that objectives for explanation methods can be stated and discussed and each explanation method can be evaluated against these objectives.
This can also give context to tasks, models or settings where some explanation methods might be suitable or unsuitable.
We will discuss later why axioms might be the most objective evaluation of explanation methods.
In the following, we briefly discuss some important axioms and their intuition.\footnote{An extensive discussion of axioms and explanation methods was done in the Individual Module and will thus not be repeated.}

\textbf{Completeness} \emph{The sum of the relevances of features of an input is equal to the prediction \citep{bach2015pixel}.}
This axiom stems from the intuition that a prediction of an input is a prediction of a composition of features. 

\textbf{Implementation Invariance} \emph{Two neural networks that are functionally equivalent, i.e.~give the same output for all possible inputs, should receive the same relevances for every input \citep{sundararajan2017axiomatic}.}
This seems trivial but is important to state as methods that work with internal weights of neural models do not necessarily comply. 

\textbf{Linearity} \emph{A network, which is a linear combination of other networks, should have explanations which are the same linear combination of the original networks explanations \citep{sundararajan2017axiomatic}.}
This axiom states that an explanation method should be a linear function of neural networks given the input.

\textbf{Sensitivity} \emph{An input feature should receive a non-zero relevance, if and only if the prediction of the network depends on the feature \citep{sundararajan2017axiomatic}.}
This is a simple double check that features that are ignored by the network should not receive relevance and features that are important to the model should receive relevance.

\textbf{Sensitivity-1} \emph{The relevance of an input variable should be the difference of prediction when the input variable is removed \citep{ancona2017towards}.}
This could also be named the occlusion axiom.
It basically restates the prediction difference formula \citep{robnik2008explaining} which we will see later in Eq. (\ref{eqn:occlusion}).

We introduce a new axiom:

\insertartificialtable

\textbf{Class Zero-Sum} \emph{The sum of relevances of a feature over all classes is zero \citep{harbecke2020considering}.}
\label{def:class_zero_sum}
The intuition behind this axiom is guided by the normalization of most classifiers.
If the sum of the predictions is fixed, then every feature adds as much to the prediction of a set of classes as it takes away from all other classes.
Thus, if we want to explain the feature for all classes we should fix the sum over all classes to zero.
E.g., we do not want a token to contribute positively to positive and negative sentiment (see Figure \ref{fig:axiom_examples}).

If a relevance method fulfills this axiom, it allows for an intuitive interpretation of the relevances, that may be taken for granted.
Since the relevances of a feature are normalized to have a sum of zero over all classes, a feature with positive relevance to a class can be interpreted as contributing to that class, whereas a feature with a negative relevance detracts from that class.

The \emph{Class Zero-Sum} axiom is a contradiction to the \emph{Completeness} axiom if the normalization of the classifier is not set to have a sum of zero.
We argue that \emph{Completeness} forces a method to assign undue relevance, e.g.~in cases where there is no information detected in the input for any class, it does not make sense to distribute positive relevance over features.

\section{Gradient-based Explanation Methods in NLP}
\label{sec:em_nlp}
Gradient-based explanation methods were introduced with Sensitivity Analysis \citep{baehrens2010explain, simonyan2013deep}.
The intuition is that the gradient of the prediction at the input tells us which input dimensions can change the prediction the most and are thus most important for the prediction.
As \cite{sundararajan2017axiomatic} showed, this can be misleading, as this only gives information about the function in the local neighborhood of the input.

In the following we call the distribution of data $p_{data}$, meaning $p_{data}(x)$ is the probability of $x$ appearing as a data point for a specific task.
This is not constrained to a dataset but the evasive more general distribution, i.e.~\enquote{data that users would expect the systems to work well on} \citep{gorman2019we}.
As discussed in section \fullref{sec:training}, in general, an input $x$ in this case can be regarded as coordinate vector over $\mathbb{R}$.
In practice this is almost always the case in NLP.
Let us assume we have a practical limitation on the length of input text and, therefore, a maximum embedding dimension of $n$.
We can make all other inputs have this dimensionality by padding them with zeros.
Thus, we have a function $f$ that maps text to a subset $S$ of a coordinate space $\mathbb{R}^n$.
We will now discuss which properties this subset $S$ has and why this is important.

\begin{figure}[ht]
  \centering
  \includegraphics[trim=0 0 0 0, width=0.9\textwidth]{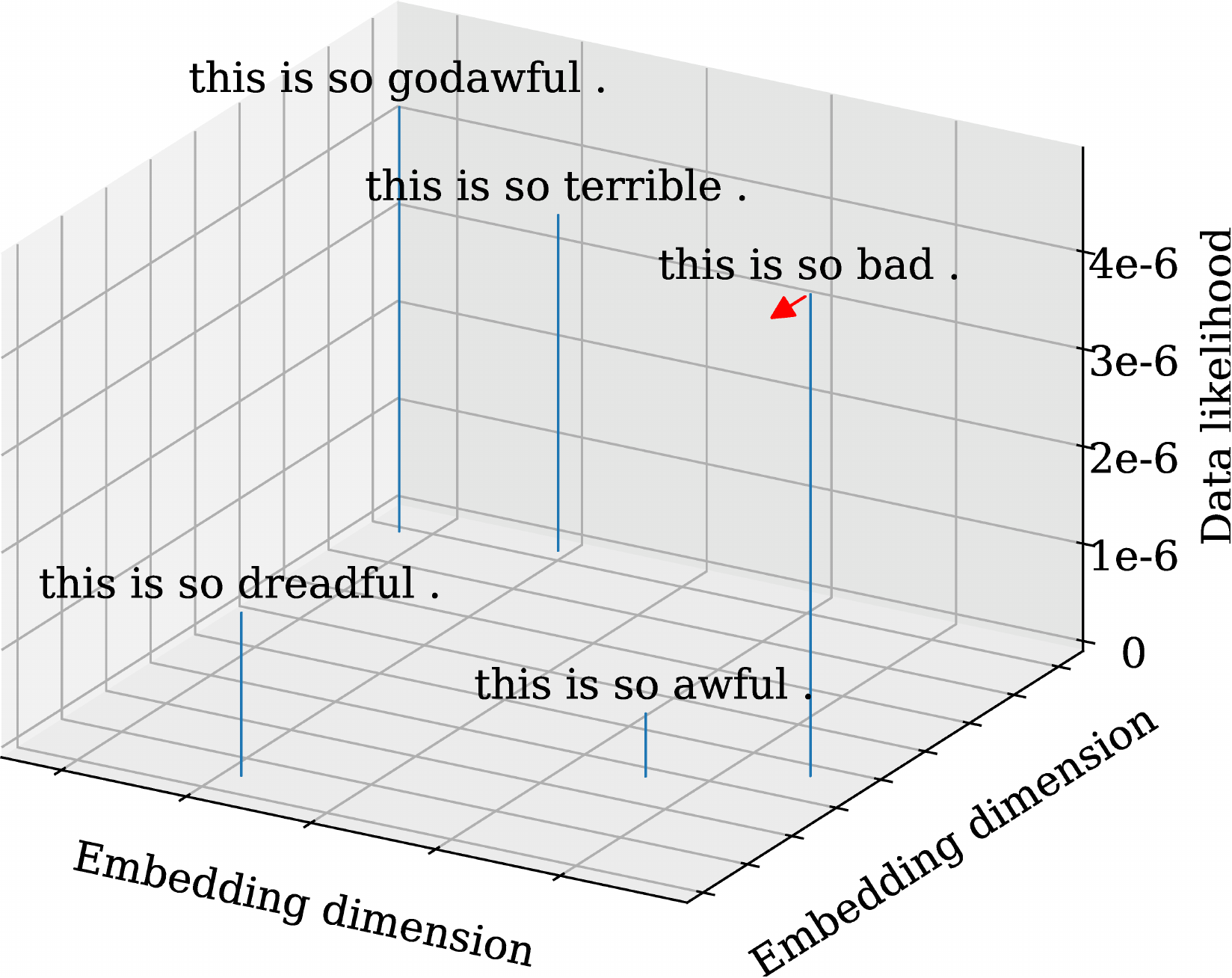}
  \caption[Schematic display of data likelihood in NLP]
  {Schematic display of data likelihood in NLP.
  There are discrete inputs, i.e., combination of tokens, with a data likelihood greater than zero.
  All other inputs in the embedding space have likelihood zero because they have no corresponding tokens.
  Every input with a positive likelihood has a neighborhood that does not contain another input with positive likelihood.
  Gradient-based explanation methods (red arrow) consider infinitesimal changes to the input and thus data with no likelihood.}
  \label{fig:likelihood_figure}
\end{figure}

\begin{enumerate}[label=\textbf{\arabic*.}, align=left, leftmargin=0pt, labelindent=0pt, listparindent=0pt, labelwidth=!]
    \item $S$ is a discrete set in $\mathbb{R}^n$. \\
    A discrete set is a set where every element $s \in S$ has a neighbourhood that does not contain any other point of $S$.
    There are a finite number of tokens with pairwise distinct embeddings and a finite length.
    Thus, $S$ is finite and the global minimum of distances between two elements of $S$ is positive.
    Therefore, $S$ is discrete.
    \item $p_{data}(x)$ for $x \in \mathbb{R}^n$ is a discrete probability distribution. \\
    For every point $x \in \mathbb{R}^n / S$ we have $p_{data}(x) = 0$.
    We just saw that $S \subset \mathbb{R}^n$ is a discrete subset.
    Therefore, $p_{data}$ is a discrete probability distribution, as can be seen schematically in Figure \ref{fig:likelihood_figure}.
    Note that this is particularly different to vision.
    Every canonical image representation $x$ in a coordinate space $\mathbb{R}^n$ entails a neighbourhood of small perturbations where the images still make sense.
    Consequently, the probability distribution of images embedded in $\mathbb{R}^n$ is continuous.
    \item This property of the probability distribution is fundamental to gradient-based explanation methods. \\
    Gradient-based explanation methods analyze the change of prediction with respect to the the input dimensions.
    Analyzing infinitesimal change in a function presumes that this change is meaningful.
    However, if the data probability is zero everywhere in a small neighbourhood of an input vector, these changes become meaningless, as the prediction function can never be confronted with other vectors from this neighbourhood.
    These vectors are automatically out of distribution, i.e.~the prediction function is analyzed by evaluating a priori meaningless behaviour.
\end{enumerate}

We thus argue that gradient-based explanation methods are not theoretically justified in NLP.
Although, in some cases, especially if the function is well-regularized, local behaviour indicates global behaviour.
This, however, can not be assumed and needs to be investigated before using gradient methods.

\section{On the Incompleteness of Evaluating Explanations}
\label{sec:explanation_evaluation}
The sole focus of an explanation method should be to relay information about a model to a user.
Explanations exist to help understand the decision process of the model \citep{doshi2017accountability}.
Thus, the correct explanation cannot be independent of the model.

Every experimental evaluation of explanation methods relies on a ground truth (property) that the explanation should have.
For example, let us take the evaluation method where the explanation of a model is compared with some features that are identified with the help of experts \citep{mohseni2018human}.
These ground truths are independent of the model.
First, neural models have achieved superhuman performance on several tasks, e.g.~go \citep{silver2016mastering} and chess \citep{silver2018general}, and grounding explanations by humans on these tasks cannot be considered helpful in understanding superhuman performance.

Second, it does not take into consideration that a model can be wrong and the explanations correct.
The explanations of a model that made a false classification can point to completely different features than an expert would select.

Third, a rigid scheme by humans does not take into account that predictive features can be missed even by experts, as neural models are powerful statistical learners \citep{sarle1994neural, geirhos2018generalisation}.
On the contrary, explanation methods are especially useful in cases where the model uses artefacts, not human-intuitive features, for its decision.

It is unclear whether it is possible to distinguish between the theoretical foundations of explanation methods and their evaluations.
The most prominent example of a method that is used for both is occlusion, which is discussed in the following section \fullref{sec:occlusion}.
Since the model is the only ground truth and explanations are a simplification, it is highly probable that there is more than one sensible explanation for an input classified by a model.
Thus, for both explanations and evaluation, constraints are established that reduce the number of explanations.
If these constrains are explicitly stated they can be regarded as axioms.
Note that axioms can be used both to develop and to test explanation methods.
All in all, the need for somewhat subjective constraints makes the existence of a general evaluation of explanation methods unlikely.

We do not argue that it is impossible to evaluate an explanation method.
The performance in sensibly selected evaluations does probably correlate to the quality of an explanation method.
Measuring the correlation to sensible explanation methods can also be seen as quality assessment.
We will use and discuss this approach in section \fullref{sec:correlation_tasks}.
Asserting which axioms an explanation method fulfills is an important step towards evaluating its validity.
We argue that the explanation method for one's use case should be motivated by the paradigms that the method fulfills.
Furthermore, there are sanity checks \citep{adebayo2018sanity} that can determine whether an explanation method has undesired properties.
Experimental evaluation of a method can provide guidance for selecting an explanation method.
They become more valuable the closer the setting of the evaluation is to the setting where the explanation is needed.

\chapter{Methods}
\label{ch:Methods}

To build a theoretically solid explanation method we combine a technique related to the \emph{Sensitivity-1} axiom with a method that considers likelihood in NLP.
We introduce these techniques, \emph{occlusion} and \emph{language modeling}, before synthesizing them to a new method.

\section{Occlusion}
\label{sec:occlusion}
We lay out the theory of occlusion and discuss its advantages and disadvantages.
Occlusion was introduced under the name \emph{Occlusion Sensitivity} \citep{zeiler2014visualizing} as a method to detect whether an explanation method detects important features of an input for a neural network, by evaluating inputs with occluded features detected by the explanation method.

Conversely, \cite{robnik2008explaining} introduce the same technique as a local post-hoc input space explanation method.
It measures the importance of features of an input for a classifier.
This is done by comparing the predictions of the classifier with and without the feature.
We refer to this method as occlusion, too.
A feature can be a word or sentence in NLP classifying or a pixel or larger patch of an image in image classifying.
Occlusion is a true black-box method.
It only uses the predictions of a model, no internal representations or even structural information about the model.

For an input $x$ we take a feature $x_i$.
To determine the relevance of $x_i$ we consider the input $x_{\setminus i}$ without this feature.
This is an incomplete input as we do not know which values to set as replacement for $x_i$.
A simple approach is to set all the values of $x_i$ to zero.
In NLP, it is possible to delete words or replace them with the \emph{<UNK>} token.
This will be done as baseline methods in the experiments.
\cite{zintgraf2017visualizing} propose sampling the values from the overall distribution of the dataset.
With this replacement we employ the difference of probabilities formula \citep{robnik2008explaining, zintgraf2017visualizing}.

The relevance $r$ given the prediction function $f$ and class $c$ is
\begin{equation}
    r_{f, c}(x_i) = f_c(x) - f_c(x_{\setminus i}).
    \label{eqn:occlusion}
\end{equation}

There are various practical difficulties with occlusion.
It is unclear which features to select, especially if the human understanding of features differs from the feature space that was used to input the data.
Furthermore, occlusion does not guarantee that the data still makes sense (to the model) after a part of the input is taken out.
This is especially difficult in NLP, where models have increasing syntactic, hierarchical and other linguistic understanding \citep{liu2019linguistic, hewitt2019structural}.
Thus, models could misinterpret data with missing features in various ways, e.g.~taking ungrammaticality as an indicator for the prediction.

\section{Language Modeling}
\label{sec:Language_Modeling}

We offer a short introduction to language modeling.
This is both valuable for our approach and for understanding the foundation of current state-of-the-art NLP classification models.
Language modeling can be seen as using large corpora of unannotated language data and creating a supervised task by reusing words as their own labels.

A DNN takes a static word embedding like \emph{word2vec} or \emph{GloVe}, or a sub-word embedding like \emph{SentencePiece} \citep{kudo2018sentencepiece} as input and predicts one or several tokens that are missing.
These missing tokens can be following the original input or masked among the input.
The labels of these tokens are one-hot vectors.
In a way, these models assign likelihood to a given text \citep{brown1992estimate}.
Let us have a text $T$ that is split into tokens $(t_1, \ldots, t_n) = T$.
If we have a language model $p_{LM}$ that is able to make predictions of the form $p_{LM}(t_1)$ and $p_{LM}(t_{i+1}|(t_1, \ldots, t_i))$ then we get a probability and score of the whole text.
\begin{equation}
    \begin{aligned}
        p_{LM}(T) & = p_{LM}((t_1, \ldots, t_n)) = p_{LM}(t_1)\prod_{i=1}^{n-1} p_{LM}(t_{i+1}|(t_1, \ldots, t_i)) \\
        PP_{p_{LM}}(T) & := p_{LM}(T)^{-\frac{1}{n}}
    \end{aligned}
\end{equation}

$PP_{p_{LM}}(T)$ is called perplexity of the corpus.
The lower this score, the higher the probability that the language model assigned to the corpus and thus, through a Bayesian argument, the better the language model.

Not all language models are trained to predict text from scratch.
E.g., \cite{devlin2019bert} mask $15\%$ of words in a sentence and predict those.
This does not lead to a model that can be measured by giving the perplexity of a corpus.

Nevertheless, all these models create an embedding of the input in every hidden layer that can be used for other tasks.
It is a priori unclear whether these representations are an improvement on the static embedding because the primary goal of the DNN is not to create a better representation of the input for other classification tasks.
However, experimental results from \cite{howard2018universal}, \cite{peters2018deep} and \cite{devlin2019bert} suggest that using embeddings from a language model is an improvement on static embeddings.
The success of these models in various tasks mentioned in section \fullref{sec:nlp} has been coined \enquote{NLP's ImageNet moment} \citep{ruder2018nlpimagenet}.
\cite{devlin2019bert} argue \blockquote{Recent empirical improvements due to transfer learning with language models have demonstrated that rich, unsupervised pre-training is an integral part of many language understanding systems.}

In our experiments, we only use language models for the simple task of predicting one missing word or punctuation mark.
This is an area where masked language models should excel.

\section{OLM}

The main idea of this thesis is combining \emph{\textbf{O}cclusion} with \emph{\textbf{L}anguage \textbf{M}odeling} (\textbf{OLM}).
Instead of leaving out tokens, we want to replace them with sensible options that only remove information but not structure.
This disallows the prediction network to consider a changed structure of the input.
It is now forced to consider the change in information.
To get a good overview of this change we sample plenty of possible replacements and compute a weighted average of the prediction results.
Due to the law of large numbers \citep{bernoulli1713ars} we consider this average an accurate approximation of all possibilities.

\subsection{Motivation}
\label{sec:olm_motivation}
Counterfactual thinking is a psychological concept which states that humans think of past and future events by looking at alternatives \enquote{What if [\ldots] ?} \citep{kahneman1981simulation, roese1997counterfactual}.
This enables us to evaluate past actions in another way than just looking at the outcome.
For statistical models we are able to actually evaluate counterfactuals without needing to guess about the outcome.
Occlusion can be seen as an objective measure to do this by changing small parts of the input and considering \enquote{What if} this part of the input is different.

These approaches are closely linked to perturbation-based explanation methods.
In contrast, these methods do not presume a set of features of which at least one is occluded, but figure out which features create the largest change in prediction if they are missing or different \citep{fong2017interpretable, wachter2017counterfactual}.
They highlight the synergy of local adversarials and explanations.
\cite{ribeiro2016should} propose a variant of this by using small perturbations to learn a local linear model that resembles the original model and is interpretable.
We argue that all these methods fail to consider data likelihood.
Furthermore, \cite{ilyas2019adversarial} show that uninterpretable adversarials are an inherent feature of almost all neural models, which questions their usefulness as explanations. 

Norm theory \citep{kahneman1986norm} states that we choose counterfactuals depending on how easy they are to imagine. 
This is at least related to how likely these alternatives are.
Rational imagination theory \citep{byrne2007rational} argues explicitly that we choose probable alternatives to reality when evaluating outcomes.
These alternatives can either lead to more positive or negative results \citep{roese2014might}.
We consider these psychological intuitions because they are important to our own understanding of explanations. 
Although, as we argued in section \fullref{sec:explanation_evaluation}, explanations should not be evaluated by human intuition, it should be clear how the results of the methods are to be understood.
Furthermore, intuitions are closely related to axioms.
The mentioned psychological statements describe intuitive principles.
The evaluation of likely alternatives is easily understood and communicable to laymen of DNNs who may examine explanations to judge a model's decision process.

\subsection{Formula Derivation OLM}
\label{sec:formula}
We start from the difference of probabilities formula in Eq. (\ref{eqn:occlusion}).
We reinterpret $x_{\setminus i}$ in the following way.
Instead of considering removing the feature $x_i$ we rather remove the information provided by this feature to the model.
This gives a very intuitive information-theoretic question that our method answers:
What additional information does this feature give to the model for classification?
For this we have to consider what inputs are how likely given the rest of the input is preserved.
In the abstract this gives
\begin{equation}
    f_c(x_{\setminus i}) = \sum_{\hat{x}_i} p_{data}(\hat{x}_i|x_{\setminus i}) f_c(x_{\setminus i}, \hat{x}_i)
    \label{eqn:p_data}
\end{equation}
with $p_{data}$ being the data probability over a defined space containing the inputs, as defined in section \fullref{sec:em_nlp}.

To use this formula in NLP we approximate the data distribution ($p_{data}$) with a language model
\begin{equation}
    p_{data}(\hat{x}_i|x_{\setminus i}) \approx p_{LM}(\hat{x}_i|x_{\setminus i}).
    \label{eqn:data_lm_approxmation}
\end{equation}

In practice this means we mask a word, which may consist of several tokens, or a punctuation mark and resample it with a given language model.
Other inputs and features are possible but could increase the approximation error, e.g.~by accumulating it over several words.
The language model does not have the information of the original word but all information of the context.
This can lead to cases where the original word is predicted with a very high probability.
We argue that in these cases the additional information provided by this word was negligent.

The big advantage of this approach is that the structure of the original input is preserved by only picking replacements that seem likely to a language model.
It gives the formula
\begin{equation}
    f_c(x_{\setminus i}) :\approx \sum_{\hat{x}_i} p_{LM}(\hat{x}_i|x_{\setminus i}) f_c(x_{\setminus i}, \hat{x}_i)
    \label{eqn:approx_missing_feature}
\end{equation}
for $x_{\setminus i}$.
There are several practical difficulties that this approximation brings.
First, it is only valid if all data is supposed to be grammatical.
A language model assigns higher probability to replacement tokens that produce a grammatical sentence.
We will investigate this effect with the \emph{Corpus of Linguistic Acceptability} dataset, in section \fullref{sec:cola_exp}.

Second, replacements sometimes make much more sense if another part of the input is changed.
For example, a prepositional verb determines the preposition it appears alongside with and the preposition can be seen as information of the verb.
This preposition, however, selects which verbs can be sampled as replacement.

Third, the length of the replacement is a limiting factor.
There are cases when the replacement should be allowed to contain more or fewer tokens than the original.
Additionally, some masked language models, which predict sub-word tokens frequently, may not sample a whole word.
This distorts the resampled input.
The last two problems can be interpreted as not giving the language model enough freedom by forcing it to make exactly one token replacement.
We elaborate possible alleviations of these difficulties in section \fullref{sec:future_work}.
In the following, we refer to the unit we want to resample as token because, disregarding practical concerns, it is arbitrary in our method.

Combined with Eq. (\ref{eqn:occlusion}) we set
\begin{equation}
    r_{f, c}(x_i) := f_c(x) - \sum_{\hat{x}_i} p_{LM}(\hat{x}_i|x_{\setminus i}) f_c(x_{\setminus i}, \hat{x}_i).
    \label{eqn:olm}
\end{equation}
This establishes a new method that can determine what effect tokens have on a prediction.
They can either have a positive or negative relevance, depending on whether the original prediction is greater than the averaged prediction after resampling.

\subsection{Axiomatic Analysis}
We will show that \emph{OLM} satisfies \emph{Class Zero-Sum}, \emph{Implementation Invariance}, \emph{Sensitivity-1} and \emph{Linearity}.
Let $f$ be a neural network that takes an element of the input space $X$ and predicts a probability distribution over classes $C$, i.e,
\begin{equation}
    \begin{aligned}
        f &: X \rightarrow \mathbb{R}^{|C|} \\
        f_c(x) & \geq 0 && \quad \forall c \in C, x \in X \\
        \sum_{c\in C}f_c(x) & = 1 && \quad \forall x \in X.
    \end{aligned}
    \label{eqn:prob_dist}
\end{equation}

Let $x_i$ be an indexed feature of an input $x$. 
We denote the relevance given to this feature regarding model $f$ and class $c$ by our method \emph{OLM} with $r_{f,c}(x_i)$.

\textbf{OLM} satisfies \textbf{Class Zero-Sum}. 
Intuitively, if the input with the resampled token increases the prediction for one class, it has to decrease the predictions of other classes, and vice-versa.
We have 
\begin{equation}
    \begin{aligned}
        \sum_{c\in C}r_{f,c}(x_i) \overset{(\ref{eqn:olm})}{=}& \sum_{c\in C} \left(f_c(x) - \sum_{\hat{x}_i} p_{LM}(\hat{x}_i|x_{\setminus i}) f_c(x_{\setminus i}, \hat{x}_i)\right) \\
        = & \sum_{c\in C} f_c(x) - \sum_{\hat{x}_i} p_{LM}(\hat{x}_i|x_{\setminus i}) \sum_{c\in C} f_c(x_{\setminus i}, \hat{x}_i)\\
        \overset{(\ref{eqn:prob_dist})}{=} &\; 1 - \sum_{\hat{x}_i} p_{LM}(\hat{x}_i|x_{\setminus i}) = 0.
    \end{aligned}
    \label{eqn:class_zero_sum}
\end{equation}

Thus, \emph{OLM} satisfies \emph{Class Zero-Sum}.
From this follows that it does not satisfy \emph{Completeness}.

\textbf{OLM} satisfies \textbf{Implementation Invariance}.
\emph{OLM} is a black-box method and only evaluates the function of the neural network.
It does not regard the parameters $\theta$.
Assume we have 
\begin{equation}
    \begin{aligned}
        \theta & \neq \theta' && \text{and} \\
        f_{\theta}(x) & = f_{\theta'}(x) && \forall x \in X. \\
    \end{aligned}
\end{equation}

Then we get
\begin{equation}
    \begin{aligned}
        r_{f_{\theta},c}(x_i) & = {f_{\theta}}_c(x) - \sum_{\hat{x}_i} p_{LM}(\hat{x}_i|x_{\setminus i}) {f_{\theta}}_c(x_{\setminus i}, \hat{x}_i)\\
        & = {f_{\theta'}}_c(x) - \sum_{\hat{x}_i} p_{LM}(\hat{x}_i|x_{\setminus i}) {f_{\theta'}}_c(x_{\setminus i}, \hat{x}_i)\\
        & = r_{f_{\theta'},c}(x_i).
    \end{aligned}
\end{equation}

Thus, \emph{OLM} satisfies \emph{Implementation Invariance}.

\textbf{OLM} satisfies \textbf{Sensitivity-1}.
\emph{OLM} is defined as an occlusion method, so it necessarily provides the difference of prediction when an input variable is occluded.
Equation (\ref{eqn:olm}) is based on Eq. (\ref{eqn:occlusion}).

\textbf{OLM} satisfies \textbf{Linearity}.
Let $f = \sum_{j=1}^{n} \alpha_j g^j$ be a linear combination of models. Then we have
\begin{equation}
    \begin{split}
        r_{f,c}(x_i) =& f_c(x) - \sum_{\hat{x}_i} p_{LM}(\hat{x}_i|x_{\setminus i}) f_c(x_{\setminus i}, \hat{x}_i) \\
        =& \sum_{j=1}^n \alpha_j g_c^j(x) - \sum_{\hat{x}_i} p_{LM}(\hat{x}_i|x_{\setminus i}) \sum_{j=1}^n \alpha_j g_c^j(x_{\setminus i}, \hat{x}_i)\\
        =& \sum_{j=1}^n \alpha_j r_{g^j,c}(x_i).
    \end{split}
\end{equation}

\subsection{OLM-S(ensitivity)}

It can be of additional interest to determine to which input features the model is most sensitive.
Previously, we measured the mean difference between the model prediction and the resampled predictions.
As a measure for sensitivity we suggest taking the standard deviation of the resampled predictions.
This measures how varied the predictions are for one token position, given the rest of the input but regardless of the original token.
With previous notation, we suggest for sensitivity $s$:
\begin{equation}
    s_{f, c}(x_i) := \sqrt{\sum_{\hat{x}_i} p_{LM}(\hat{x}_i|x_{\setminus i}) \left(f_c(x_{\setminus i}, \hat{x}_i) - \mu \right)^2}.
\end{equation}

We do not suggest this as a relevance measure because, as previously mentioned, it is independent of the input feature $x_i$.
Rather, this measure suggests additional information to the relevance method.
For a neutral feature it may be reassuring to know whether the model would have picked up on more class-indicating possibilities.
In combination \emph{OLM} and \emph{OLM-S} measure the mean and standard deviation of predictions with resampled tokens.

\chapter{Experiments}
\label{ch:Experiments}

We perform several experiments to investigate the explanations generated by \emph{OLM}.
Example explanations can be found in Table \ref{tab:example_explanations}.
These experiments highlight some practical peculiarities of our explanation method.
For comparison with other methods we also display the results of \emph{OLM-S}.
These experiments cannot be comprehensive (see section \fullref{sec:explanation_evaluation}) and there is no standard benchmark.
Thus, we compare to other explanation methods and conduct experiments in areas that may present corner cases to \emph{OLM}.
\insertexampletable

Underlying these experiments is mostly the same combination of explanation and prediction model.
This is done for two reasons.
We investigate the algorithmic efficiency of \emph{OLM}.
For every token we resample $k$ times.
Let us assume the distribution of these tokens follows a variation of Zipf's Law \citep{estoup1916gammes, zipf1949human} with $\alpha > 1$ \citep{piantadosi2014zipf}.
Then, we have $O(\sqrt[\alpha]{k})$ different samples per token.
Furthermore, we resample each of $n$ tokens in an input.
Thus, for a single input we have $O(n \sqrt[\alpha]{k})$ predictions.
An investigation of the effect of different language models on the explanations of different classification models should be done but requires vast resources.

Additionally, we fix the models to compare results across different tasks and datasets.
To this end, we also only investigate explanations of the true label neuron.
Some explanation methods do not necessarily treat different classes differently, as alluded to in Figure \ref{fig:axiom_examples}.
We try to remove this effect by focusing only on the most important class.
\insertsamplingtable

For \emph{OLM} and \emph{OLM-S} we use $\textsc{BERT}_\textsc{base}$ \citep{devlin2019bert} as a language model and choose words (and punctuation marks) as units for resampling.
Resampling is computationally expensive but the quality of the samples is very important.
We also want a language model that does not frequently produce sub-word tokens.
\textsc{BERT} uses \emph{WordPiece} \citep{wu2016google} which does have sub-word tokens but mostly predicts whole words.
This is viable for single word resampling, especially compared to many other masked language models.
Thus, we choose $\textsc{BERT}_\textsc{base}$ as a low-resource compromise of a well-fitting state-of-the-art language model to analyze the method over datasets.
An example of the samples is shown in Table \ref{tab:sampling_examples}.
We point out that in general our approach is language model agnostic.
For generating single input explanations, not analyzing a dataset, we suggest using a collection of the best well-fitting language models available.
For classification we use different variations of \textsc{RoBERTa} \citep{liu2019roberta}\footnote{All models were originally published at \url{https://github.com/pytorch/fairseq/tree/master/examples/roberta}. We use the implementation and pre-trained models from \url{https://github.com/huggingface/transformers}.} that we describe in section \fullref{sec:correlation_tasks}.\footnote{Experiments are available at \url{https://github.com/harbecke/xbert}.}

\section{Correlation of Explanation Methods}
\label{sec:correlation}

First, we compare the relevances produced by \emph{OLM} to those of other explanation methods.
The main focus of this experiment is to evaluate how large the differences to other explanations are.
It could be assumed that the explanations of basic occlusion is very similar to \emph{OLM} explanations.
If this were shown by experiments, it would make the theoretical benefits of \emph{OLM} superfluous.
E.g., in theory, the language model of a state-of-the-art classifier could understand an obviously missing word.
It could treat this similar to our method by internally representing it as missing and deriving a prediction from that.
In the same vein, we investigate how much the explanation methods that use gradients differ from occlusion-based methods to see if theoretical difficulties manifest.
\footnote{
This experiment already appears in \cite{harbecke2020considering}.
The baseline methods were selected by me.
Christoph Alt selected and conducted the experiments.
Phrasing and analysis exceeding the publication is mine.
}

We calculate the correlation of explanation methods on tasks in the following way.
Let $r^{1}_{x}$ and $r^{2}_{x}$ be the ordered relevances of an element $x\in X$ of dataset $X$ for methods $1$ and $2$.
With $\textbf{corr}$ being the Pearson correlation coefficient for samples \citep{pearson1895vii}, we set the correlation of two methods over dataset $X$ to
\begin{equation}
    \frac{\sum_{i=1}^{n} \operatorname {corr} (r_i^1, r_i^2)}{n}.
\end{equation}
This means, two methods are perfectly positively correlated if and only if they produce scaled relevances with possibly different positive scaling for each input.

\subsection{Baseline Methods}

We compare our explanations with two baselines based on occlusion (see Eq.~(\ref{eqn:occlusion})).
The simplest variation is removing the word of interest and not replacing it.
We call this method \textbf{Delete}, it was first used in NLP by \cite{li2016visualizing}.
Similarly, we replace the word of interest with the unknown token <\textbf{UNK}>.
This approach is more tailored to state-of-the-art classifiers pre-trained with masked language modeling.

Furthermore, we compare explanations to three gradient-based methods.
All these methods provide relevances for every dimension of the input.
To receive relevances on word level we sum over the dimensions for each word \citep{arras2017relevant}.
The simplest one is the absolute value of the gradients and is called \textbf{Sensitivity Analysis} \citep{simonyan2013deep}.
Note that this method only provides non-negative relevances.
It is especially comparable to \emph{OLM-S} which also provides a non-negative sensitivity of the model.
\textbf{Input*Gradient} \citep{shrikumar2017learning} is self-explaining.
Every input gets multiplied with its gradient.
Finally, we compare our explanations to \textbf{Integrated Gradients} \citep{sundararajan2017axiomatic}.
This is the integration of the gradient of the prediction function along a straight path from a baseline, usually the zero vector, to the input vector multiplied by the path length.

\subsection{Tasks}
\label{sec:correlation_tasks}

We select three NLP classification tasks.
An input contains one or two sentences or phrases for all tasks.
Each task focuses on one specific aspect of language understanding.
All tasks are part of the GLUE benchmark \citep{wang2019glue} which does not publish test sets.
Therefore, we report results on the development set which we do not use for model optimization.

\textbf{Multi-Genre Natural Language Inference Corpus (MNLI)} by \cite{williams2018broad} is a natural language inference corpus.
A data point consists of two sentences that may have a relation to each other.
\begin{itemize}
    \item If the second sentence is a sensible successor to the first in content, this pair gets the \emph{entailment} label.
    \item If the content of the sentences does not relate to each other, the label is \emph{neutral}.
    \item If the sentences have are in disagreement they get a \emph{contradiction} label.
\end{itemize}
The corpus contains more than 400,000 samples.

We use a $\textsc{RoBERTa}_{\textsc{LARGE}}$ model which is already fine-tuned on MNLI.
It achieves an accuracy of $90.2\%$ on the development set which is two percentage points behind state-of-the-art \textsc{T5} \citep{raffel2019exploring}.
\cite{mccoy2019right} show that even though these models perform around the human baseline for this task, they fail to generalize for a variety of rare constructions.
Correlations of the explanation methods for MNLI can be found in Table \ref{tab:mnli_method_correlation}.

\insertmnlicorrelationtable

\textbf{Stanford Sentiment Treebank} (SST) by \cite{socher2013recursive} is a sentiment classification dataset.
It contains 70,000 sentences from movies with either a negative or positive connotation.
The SST-2 version only consists of binary classification with positive and negative sentiment.
Sentiment Analysis is an easy task to interpret explanations on if the explanation method assumes that features cannot contribute to both classes (see Figure \ref{fig:axiom_examples} and Eq.~(\ref{eqn:class_zero_sum})).
An input feature contributes as much to the positive sentiment as it detracts from the negative sentiment and vice versa.
Therefore, the explanation method assigns each feature positive or negative sentiment.

We fine-tune a pre-trained $\textsc{RoBERTa}_{\textsc{BASE}}$. 
This model achieves an accuracy of $94.5\%$ on the development set which is 3 percentage points lower than multiple state-of-the-art models, including \textsc{T5} \citep{raffel2019exploring}.
Correlations of the explanation methods for SST-2 can be found in Table \ref{tab:sst2_method_correlation}.

\insertsstcorrelationtable

\textbf{Corpus of Linguistic Acceptability} (CoLA) by \cite{warstadt2019neural} is a dataset with sentences labeled by their grammatical acceptability.
It contains more than 10,000 sentences which are annotated as either acceptable or unacceptable.
We will elaborate on the specifics of this task for explanation methods in section \fullref{sec:cola_exp}.

Analogous to SST-2 we fine-tune $\textsc{RoBERTa}_\textsc{BASE}$ and achieve a phi coefficient\footnote{also misnomered Matthews correlation coefficient} \citep{yule1912methods} of 0.613 on the development set.
\textsc{StructBERT} \citep{wang2019structbert} achieves a phi coefficient of 0.753.
Correlations of the explanation methods for CoLA can be found in Table \ref{tab:cola_method_correlation}.

\insertcolacorrelationtable

\subsection{Results}

Tables \ref{tab:mnli_method_correlation}, \ref{tab:sst2_method_correlation} and \ref{tab:cola_method_correlation} show correlation of all tested explanation methods.
Overall, there is positive correlation between almost all methods.
No methods produce equivalent relevances, even the correlation  between \emph{OLM} and \emph{OLM-S} is never close to 1.
The three occlusion-based relevance methods \emph{OLM}, \emph{Delete} and \emph{UNK} have consistently high correlation on MNLI and SST-2 but much lower correlation on CoLA.

We draw the following conclusions.
\emph{OLM} produces significantly different explanations than other occlusion methods.
\emph{Delete} and \emph{UNK} have a higher correlation with each other for all three tasks than with \emph{OLM} which is evidence that it stands out from other occlusion methods.
The theoretical differences between these methods seem to manifest experimentally.

\emph{OLM-S} has the about as much correlation to other occlusion methods than to \emph{Sensitivity Analysis}.
This can be seen as experimental validation as a sensitivity method with a somewhat different intent than a relevance method.

The correlation between gradient-based methods and other methods is low across all tasks.
This could indicate that gradient methods do not capture the discrete nature of NLP (see section \fullref{sec:em_nlp}).
Note that we use correlation in our argument in two seemingly conflicting ways.
If the correlation of explanations is close to 0, they are independent of each other.
It is unlikely that methods with independent results can both give very good explanations.
Furthermore, if the correlation of explanations is close to 1, the methods are redundant.
They might have theoretical differences but these at least did not manifest in the experiments.
On the whole, we think different sensible explanation methods should have positive correlation $0 \ll c \ll 1$.
This appears to be the case for \emph{OLM} and most compared explanation methods.

Table \ref{tab:example_explanations} shows example explanations for the input \enquote{good film , but very glum .} from the SST-2 dataset.
Table \ref{tab:sampling_examples} shows the sampling examples that \emph{OLM} used to give relevance to the words.
The maximum value is easy to interpret for occlusion-based methods, as it indicates the change in prediction if the feature with the maximum value is occluded.
\emph{OLM} and \emph{OLM-S} give lower relevances to \enquote{good} because alternatives sometimes lead to a positive classification as well.
Interestingly, alternatives to \enquote{film} more often lead to a negative classification.
This indicates that the meaning or strength of \enquote{good} depends on the following noun.
Surprisingly, \enquote{glum} also receives a positive relevance from \emph{OLM}.
There are alternatives, especially \enquote{bad}, that change the prediction to negative.
Furthermore, the classification gives positive sentiment a probability of $98\%$, so no negative word had a big effect in the original sentence.

For two out of five words the original word was the most frequent sample from the language model.
Additionally, the comma was resampled $84\%$ of the time and the period every time.
This indicates that the information of these was not completely lost as they are made very likely by the context.
Many resampled words (prediction $< 0.5$) lead to a negative sentiment classification.
Also, many resampled words lead to a classification where the model assigns a high probability to one class, i.e.~is very sure. 
Only the first word has several replacements where the model is unsure.

This example also highlights a disadvantage of \emph{OLM}.
The structure \enquote{good [...] but [...] glum} can not be adequately evaluated by resampling only one word.
It is possible to argue that two of these words already determine the sentiment of the third.

\section{CoLA Corner Case}
\label{sec:cola_exp}

CoLA is an interesting edge case for our method.
In the CoLA dataset there are inputs that are grammatically acceptable and inputs that are unacceptable.
The task of a model is to find out to which category the input belongs.
The language model in \emph{OLM} mostly saw acceptable sentences during training, thus it can be assumed that \emph{OLM} tries to resample such that grammatically acceptable input appears whenever possible.
This can be seen as flawed, because on average the resampling should lead to a similar class distribution as the original dataset.

Therefore, in theory, \emph{OLM} is able to identify important words if replacements make the sentence acceptable, because it assigns relevance if the prediction of the model changes from ungrammatical to grammatical.
However, it does not identify the opposite case, where an input is grammatical only because of a specific construction.
\emph{OLM} would not modify the input to be unacceptable.
In other words, CoLA is a task where the approximation $p_{data}\approx p_{LM}$ in Eq.~(\ref{eqn:data_lm_approxmation}) is not necessarily true, the resampling is not faithful to the data.

\subsection{Statistical Analysis of Explanations}
\label{sec:cola_statistics}

To investigate this, we try to answer the following question.
How much do the explanations for grammatically acceptable and unacceptable sentences differ?
We hypothesize that the explanations for unacceptable inputs have higher values on average.

First, we have to set the classes on equal footing.
We only use explanations of inputs that were predicted correctly by the $\textsc{RoBERTa}_{\textsc{BASE}}$ model from section \fullref{sec:correlation} and with a probability $p$ of at least $0.9$.
The relevances produced by \emph{OLM} are always in the range $[p-1, p]$, as for all occlusion-based methods (see Eq.~(\ref{eqn:occlusion})).
This ensures that all explanations fall into the range $[-0.1, 1]$.
We have $165$ sentences labeled as unacceptable and $678$ labeled as acceptable with probability $p \geq 0.9$.

We measure the sum, average and maximum of relevances over input sentences.
For each of these measures we get results for acceptable and unacceptable sentences, which can be compared with a statistical significance test.
We choose \emph{Welch's t-test} \citep{welch1947generalization} which enables samples sizes and variances to be different for both classes.
We can ignore that the samples are not from a normal distribution because of the large sample size \citep{kendall1951advanced}.
Table \ref{tab:cola_significance} shows results of these tests.
The null hypothesis of the averages being equal can be rejected for all comparisons because all three tests are highly significant.

\insertcolasignificancetable

We feel obliged to mention that this is not a conclusive test.
Features of inputs of different classes need not have equivalent properties.
For acceptability datasets, there are many more possible constructions for unacceptable sentences than acceptable sentences.
Thus, we must not expect a symmetry between the explanations of different classes.

The other two occlusion methods \emph{Delete} and \emph{UNK} find more relevance in the acceptable sentences.
This is likely due to them perturbing grammatically acceptable input such that it is not acceptable.
The \emph{Delete} method removes words, the \emph{UNK} method replaces them with the \emph{<UNK>} token, which may not share all syntactic properties with the original word.
Thus, they are both likely to break the syntax of a sentence.
This confirms that our method does differ significantly from other occlusion-based methods.

\insertcolaexampletable

We show some randomly selected examples of the explanations in Table \ref{tab:example_explanations_cola}.
The sentences correctly classified as unacceptable all have words that can be replaced to make the sentence acceptable.
The sentences correctly classified as acceptable have few words with high relevance, so the language model rarely created sentences that the classifier considered unacceptable.
Examples of these resampled sentences can be found in Table \ref{tab:resamples_cola}.

\insertcolaresamples

We take this as evidence that the language model is able to construct grammatical sentences even under adverse circumstances.
The inspection of this dataset with \emph{OLM} can be a useful analysis.
However, the approximation in Eq.~(\ref{eqn:data_lm_approxmation}) of the language model distribution approximating the data distribution does not hold up.
The relevances of words in unacceptable sentences is amplified because the language model tries to choose words that build an acceptable sentence.
We consider the results of \emph{OLM} on this task meaningful but not unconditionally.
The task shows investigating the approximation, or adapting the language model for the classification task, can be pondered.

\section{FAVA}

We conclude with an experiment that resembles CoLA but with a specific linguistic aspect and a possibility for more model introspection.
\cite{kann2019verb} introduce the Frames and Alternations of Verbs Acceptability (FAVA) dataset. 
It contains constructed sentences around verb properties that yield acceptable and unacceptable sentences.
Most of these sentences come in pairs of acceptable and unacceptable sentence, which are variations of each other and at least one uses the given verb frame.
This allows a direct comparison that exceeds the possibilities of CoLA.
We only select these pairs for evaluation.
Example sentences and frames can be found in Table \ref{tab:example_explanations_fava}.

\insertfavaexampletable

In this case we do not fine-tune a model on the dataset but use the\linebreak[4] $\textsc{RoBERTa}_\textsc{BASE}$ model from \emph{TextAttack} fine-tuned on CoLA.\footnote{The model was originally published at \url{https://github.com/QData/TextAttack}. We used the version from the transformers package \url{https://github.com/huggingface/transformers}.}
Since this model was trained on another dataset, this allows us to evaluate on the train, development and test set.
We do this to retrieve enough data.
This dataset version encompasses 6466 single sentences or 3233 pairs.
The accuracy for the model on our class-balanced version of the dataset is $84.3\%$.
The best model in the original paper \citep{kann2019verb} achieved an accuracy of $85.5\%$ on the test set of the whole dataset.

This dataset is split into five types of syntactic verb frame alternations.
Verbs can be seen as the central part-of-speech to this task, as constructions are built around the fact that they work for some verbs but not for all.
Thus, we are interested whether verbs have a higher relevance than other words.
We perform significance tests as in section \fullref{sec:cola_statistics}.
Table \ref{tab:fava_significance} shows statistical significance for both acceptable and unacceptable sentences.
For unacceptable sentences the relevance of verbs is more than three times as large as for the average word.
For acceptable sentences the relevance is more than two times as large.
This shows that \emph{OLM} identifies verbs as important words in this task which is centered on verbs.

\insertfavasignificancetable

In Table \ref{tab:example_explanations_fava} we show an example of the explanation of sentence pairs for each of the syntactic verb frame alternations.
For all verb frames the verb gets the highest relevance in at least one of the two paired sentences.
In only one of the verb frames the verb has the highest relevance in both paired sentences.
As expected, the relevances of the unacceptable sentences are higher than the relevances of acceptable sentences.

\insertfavacorrelationfigure

We also investigate how the relevances of verbs in sentence pairs are correlated.
We already saw that the relevances are higher for words of unacceptable sentences.
Figure \ref{fig:verb_correlation} provides a visualization of the relevances of sentence pairs.
The null hypothesis of the relevances not being correlated is rejected with $p$-value $<0.005$.
The relevances are weakly negatively correlated with a Pearson's correlation coefficient of $r=-0.134$.
We interpret this as a sanity check \citep{adebayo2018sanity} that our method does not simply assign high relevance to verbs.

The relevance of verbs in unacceptable sentences is higher on average, as is the case for an average of all words in the CoLA experiments.
Few verbs have high relevance in both sentences of a sentence pair.
This can have two possible reasons.
If the classification model does not contextualize a verb completely, the verb as a feature itself influences the model prediction.
This would mean an imbalance for the resampling predictions.
Also, the verbs are unlikely to be special cases of both the unacceptable and the acceptable sentence.
Resampling is more likely to change the prediction if the construction around the verb only does or does not work for few samples.

All in all, these experiments provide evidence that \emph{OLM} is a useful tool for analyzing black-box models.
At the least, it provides easily interpretable explanations.
We identify language modeling as a standout feature of \emph{OLM} that makes its explanations vastly different from other methods.
The intuitive theory of \emph{OLM} allows for evaluation and introspection of relevances even for corner cases of the method.

\chapter{Conclusion}
\label{ch:Conclusion} 

We summarize the topics and central arguments of this thesis.
Furthermore, we point out weaknesses of our approach and how they could be alleviated.
Last, we elaborate some possible future work.

\section{Summary}

In the first chapter of this thesis we provide the necessary theoretical background.
We introduce Deep Learning by motivating interest in the topic, mainly through its achieved results.
Additionally, we discuss practical aspects, such as the architectures of deep neural networks and how to train them.
We round out this section by pointing to capabilities of neural networks and how they are important to this thesis.

We introduce the topic of explainability of neural networks.
We summarize central papers and surveys of this topics.
Focus is laid on theoretical aspects of explainability.
This is done partly by showing work on axioms for explanation methods.
In this context we provide our first contribution, a novel axiom.
It is guided by the principle that a feature of an input to a normalized prediction function contributes as much to a set of classes as it detracts from the complementary classes. 

The second contribution is a theoretical argument against gradient-based explanation methods in natural language processing.
We show that input in NLP is discrete and thus the data likelihood distribution is discrete.
This inhibits the functionality of gradients when analyzing the prediction function.
We contrast this with the likelihood function in vision.

Furthermore, we contribute another theoretical argument to explainability.
We discuss why a general evaluation of explanation methods is unlikely to exist.
The main argument is that the ground truth for the explanation is only held in the model, which is exactly what an explanation method is trying to extract.
Since this can only be done by approximation, we point out that there is a false dichotomy between the evaluative rules for this approximation and the explanation method that fulfills them.

The central methods chapter introduces our main contribution, a novel explanation method.
It consists of the combination of two existing methods, \emph{\textbf{O}cclusion} and \emph{\textbf{L}anguage \textbf{M}odeling} and is coined \textbf{OLM}.
Occlusion is a technique in explainability that is used either to explain black-box models or evaluate explanations.
It can be seen as incomplete because it does not determine how to replace occluded features.
For this replacement we propose using language modeling in NLP.
We show that language modeling is especially suited to sample replacement.
They excel at this task because language understanding by language models is the foundation of state-of-the-art models in NLP.

We motivate \emph{OLM} psychologically and through information theory.
These arguments lead us to selecting and evaluating likely alternatives to explain features of the original input.
The formula for \emph{OLM} is achieved by inserting sampling with a language model into the difference of probabilities formula.
We analyze our method by going over the axioms that we introduced in the preceding chapter.
We provide proofs of compliance.
Since the formula for \emph{OLM} is a weighted mean of predictions, we add the standard deviation as an additional measure \emph{OLM-S}.
This measure is intended as a sensitivity method, not a relevance method.

We provide experimental evaluation of our method.
As alluded to in the second chapter, we do not consider a complete evaluation, like a single benchmark, possible.
We try to compare our method to existing methods by a correlation experiment.
We work under the assumption that sensible methods correlate, but when introducing a novel method it is also important to see that there is no perfect correlation.
Both aspects are shown over three tasks.
The results on one task, CoLA, point to a possible problem of \emph{OLM}.

The second experiment is a deeper dive into the relevances of our method on CoLA.
Linguistic acceptability is a context where it is wrong to assume that the distribution of a language model is similar to the distribution of the data.
We show that this manifests in the explanations.
We discuss how to interpret these explanations and how they make sense once the context is understood.
The introspection with \emph{OLM} is possibly deeper than with other relevance methods, as we can also provide the compared samples.
Furthermore, we show that the language modeling works as it is supposed to.

Finally, we experiment on a dataset that is even more specific than CoLA.
It contains pairs of sentences with verb frames that are either acceptable or unacceptable.
This provides the additional information that the verb has to be very important for the model, as different verbs lead to different classifications in the same structure.
We show that our method is able to identify verbs as very important.
Furthermore, we show that relevances of verbs in sentence pairs is weakly negatively correlated, which indicates that our method does not blindly assign relevance to verbs.

\section{Future Work}
\label{sec:future_work}

We describe some of the shortcomings, unexplored areas and possible ideas to extend this work.
All identified weaknesses concern the practical application of our method.

There are three main problems with our approach.
The major problem we identified during method development and experiments is the approximation of the data with a language model.
While this is certainly valid frequently, we saw that there are exceptions.
A language model can always provide syntactically correct data which does not make sense for the task.
Of all possible sentences, most sentences are not suitable for a dataset because they do not indicate a class.
A language model alone has no mechanism to detect this.

One possibility to improve selection of resampled tokens would be to train a generative adversarial network \citep{goodfellow2014generative} to detect whether an input is part of the given dataset.
The discriminator of this model could determine whether proposed resampled tokens create a likely input and also assign a probability.
This probability could be used or combined with the language model probability to select resampled tokens.
This would ensure that the resampled data fits the task.

A second problem of our method is that we resample exactly one token in a tokens previous place.
If we choose words as replacement units, this prevents the use of language models that produce sub-word tokens frequently.
It would be possible to use beam search to allow them to build a word from these sub-word tokens.
Additionally, If we want to change the position, we would not know where to position a replacement.
This is problematic in several ways.
In general, if we take out the information of one word then there are other sensible replacements that do not consist of just one word.
Furthermore, if the replaced word forces a syntactic structure in the input that is not determined by any of the other inputs, it could be sensible to allow this structure to change.

In an information theory sense, the removal of the information of one token places the sentence into another set of possibilities that contain the information of all other tokens but not necessarily in the exact phrasing as before.
This is the same phenomenon that humans have when thinking of formulating a sentence and changing the structure at the last moment because the last word fits the new structure better.
To allow this, we need an architecture that is invariant to rephrasing.
It would be possible to use an encoder-decoder architecture to encode the information of the sentence without the tokens of interest.
Then we have a representation of the sentence that can be used to generate the replacement sentence.
However, it would be difficult to detect whether there is information missing that needs to be sampled in the encoded state.
E.g., if a strong adjective is removed in a sentiment classification task, how do we produce a resampled sentence with sentiment?

Another interesting aspect is choosing other features as replacement units.
For tasks with many sentences it could be interesting to measure the effect of sentences as features.
However, we consider it unlikely that resampled sentences produce vastly different results than removed sentences. 
Syntax is no longer an issue, only context for other sentences may be important.

The third practical difficulty is the dependency of our approach on a language model.
Even state-of-the-art language models differ from the true likelihood of language data.
This problem is unlikely to be resolved practically, as language models will presumably not become perfect in the future.
Our method depends on automatic generation of replacements.
This complicates resampling multiple features at the same time, as the approximation errors are likely to accumulate.

All mentioned practical points concern the possibilities and difficulties of generating good comparison samples.
This thesis is an attempt towards explanation methods in natural language processing that do not use unjustified inputs or methods.
Therefore, a deeper analysis of sampling options to improve the tailoring of \emph{OLM} to NLP is an adequate future research direction.

\phantomsection
\addchaptertocentry{Bibliography}
\bibliography{thesis}

\appendix

\end{document}